\documentclass{article} % For LaTeX2e
\usepackage{conference,times}

\usepackage{amsmath,amsfonts,bm}

\def\eqref#1{equation~\ref{#1}}
\def\1{\bm{1}}

\DeclareMathAlphabet{\mathsfit}{\encodingdefault}{\sfdefault}{m}{sl}
\SetMathAlphabet{\mathsfit}{bold}{\encodingdefault}{\sfdefault}{bx}{n}

\usepackage{hyperref}
\usepackage{url}
\usepackage{amsmath,amsfonts}
\usepackage{array}
\usepackage[caption=false,font=normalsize,labelfont=sf,textfont=sf]{subfig}
\usepackage{textcomp}
\usepackage{stfloats}
\usepackage{verbatim}
\usepackage{graphicx}
\usepackage{amsmath}
\usepackage{amssymb}
\usepackage{booktabs}
\usepackage{color}
\usepackage{multirow}
\usepackage{enumerate}
\usepackage{bbding}
\usepackage{wrapfig}

\newcommand{\mk}[1]{\textcolor{blue}{#1}}
\newcommand{\dlt}[1]{}

\title{UC-NeRF: Neural Radiance Field for Under-Calibrated multi-view cameras in autonomous driving}

\author{Kai Cheng$^1$, Xiaoxiao Long$^2$\thanks{First two authors contributed equally.}, Wei Yin$^3$, Jin Wang$^1$, Zhiqiang Wu$^2$, Yuexin Ma$^4$, \\ \textbf{Kaixuan Wang}$^3$, \textbf{Xiaozhi Chen$^3$}, \textbf{Xuejin Chen$^1$\thanks{Corresponding athuor (xjchen99@ustc.edu.cn).}} \\
$^1$ University of Science and Technology of China\\
$^2$ PKU-Wuhan Institute for Artificial Intelligence \\
$^3$ DJI Technology \\
$^4$ ShanghaiTech University\\
}

\begin{document}

\maketitle

\begin{abstract}
Multi-camera setups find widespread use across various applications, such as autonomous driving, as they greatly expand sensing capabilities. 
Despite the fast development of Neural radiance field (NeRF) techniques and their wide applications in both indoor and outdoor scenes, applying NeRF to multi-camera systems remains very challenging. This is primarily due to the inherent under-calibration issues in multi-camera setup, including inconsistent imaging effects stemming from separately calibrated image signal processing units in diverse cameras, and system errors arising from mechanical vibrations during driving that affect relative camera poses.
In this paper, we present UC-NeRF, a novel method tailored for novel view synthesis in under-calibrated multi-view camera systems.
Firstly, we propose a layer-based color correction to rectify the color inconsistency in different image regions. Second, we propose virtual warping to generate more viewpoint-diverse but color-consistent virtual views for color correction and 3D recovery. Finally, a spatiotemporally constrained pose refinement is designed for more robust and accurate pose calibration in multi-camera systems.
Our method not only achieves state-of-the-art performance of novel view synthesis in multi-camera setups, but also effectively facilitates depth estimation in large-scale outdoor scenes with the synthesized novel views. See the project page for code, data: \href{https://kcheng1021.github.io/ucnerf.github.io/}{\mk{https://kcheng1021.github.io/ucnerf.github.io/}}.

\end{abstract}

\section{Introduction}

\begin{wrapfigure}{r}{0.25\textwidth}
\vspace{-0.8cm}
\includegraphics[width=1\linewidth]{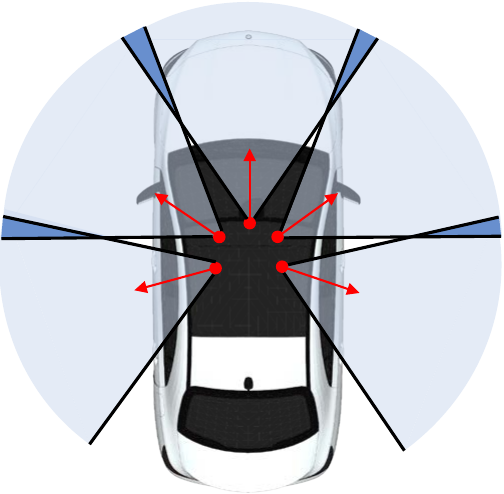}
\vspace{-0.8cm}
\label{fig:multicam}
\caption{Illustration of a multi-camera system.
% Different cameras are placed in fixed positions with relatively small overlapping field of view.
} 
\vspace{-0.5cm}
\end{wrapfigure}

Neural radiance field (NeRF) is a revolutionary approach that enables the synthesis of highly detailed and photorealistic 3D scenes from 2D images. This technology has opened up a multitude of new possibilities in autonomous driving, such as generating synthetic data from diverse viewpoints for robust training of perception models and providing effective 3D scene representations to enhance comprehensive environmental understanding (\cite{fu2022panoptic,zhang2023nerflets}).

Multi-camera systems (\cite{sun2020scalability, caesar2020nuscenes, guizilini20203d}) are commonly used for autonomous driving, while involving the strategic placement of multiple cameras to capture a holistic perspective of the surrounding environment, as shown in Fig.~\ref{fig:multicam}, supplying spatially consistent information to complement the temporal data for perception tasks (\cite{mei2022waymo, pang2023standing}). 
Incorporating NeRF in multi-camera systems could provide a way to efficiently and economically produce extensive high-quality video data for training various models in autonomous driving systems. 

\begin{figure*}[t!]
\centering    
\includegraphics[width=\linewidth]{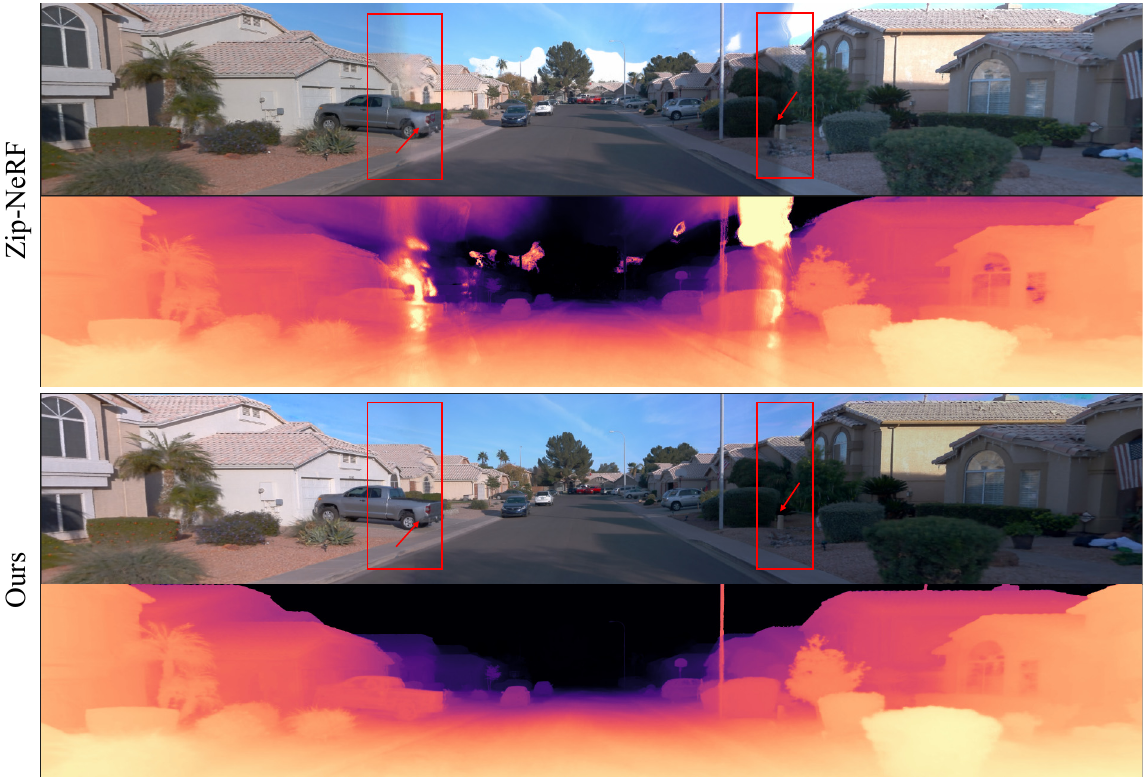}
\caption{In under-calibrated multi-camera systems,  the NeRF quality significantly degrades (the first row), along with color discrepancies (red boxes), object ghosts (red arrows), and wrong geometry (the second row). Our UC-NeRF achieves high-quality rendering and accurate geometry in the challenging cases.} 
\label{fig: motivation}
%\vspace{-0.5cm}
\end{figure*}

However, naively combining images captured from multi-camera systems into NeRF's training often results in a significant deterioration of rendering quality, as depicted in the first two rows of Fig.~\ref{fig: motivation}. The underlying cause of this degradation is the inconsistent color supervision from different views, since the multi-cameras are usually under-calibrated. This under-calibration issue manifests in two distinct ways.
First, the image signal processing (ISP) units involve various techniques like white balance correction, gamma correction, etc., to convert the raw data captured by the sensors to discretized pixel colors. However, these ISP units fluctuate over time across different cameras, resulting in color disparities within the same scene region, as shown in the red box in Fig.~\ref{fig: motivation}.
Secondly, even with delicate camera pose calibration beforehand, systematic errors during vehicle production and vibrations during driving inevitably introduce misalignment and further exacerbate color inconsistency, as indicated by the arrows in Fig.~\ref{fig: motivation}.

Several NeRF methods (\cite{martin2021nerf, rematas2022urban, tancik2022block}) attempt to alleviate the inconsistent color supervision by modeling image-dependent appearance with a global latent code for each image.
%场景不同区域对应不同的ISP effect,一个全局变换的建模能力有限；多相机之间场景的overlap比较小，对齐不同相机间包含的场景的颜色空间比单个相机要困难。
However, the capacity of a global latent code to uniformly correct colors in different regions of an image is limited, especially when different regions correspond to different color transformations. 
Furthermore,
learning one color correction for each image can lead to overfitting when the training images lack color and viewpoint diversity. This limitation is pronounced for areas observed by side-view cameras, which have fewer observations and limited overlapping with front-view areas.
% in multi-camera systems, the overlap area of scenes observed by different cameras is limited. This makes aligning the color spaces of images under different cameras more challenging than with a single camera.
% the lack of constraints on these latent codes often leads to overfitting to training views, causing a drop in rendering quality.
To correct inaccurate poses, some approaches perform joint NeRF optimization with pose refinement using photometric losses. Unfortunately, utilizing such joint optimization under a multi-camera setup,
% all parameters become highly correlated, 
the photometric consistency across cameras can not be ensured and spatial relations among cameras are not fully utilized, making optimization more challenging and prone to local minima.
% In contrast, our approach decouples pose optimization from NeRF optimization and explicitly incorporates a spatial-temporal constraint among cameras. This prevents under-constrained optimization and yields more accurate results.
% but most of them ignores the spatial constraints among the multiple cameras
% Some other NeRF methods implement joint pose optimization, but they heavily rely on photometric losses, making it challenging to achieve precise pose optimization when the images captured by different cameras inherently exhibit photometric inconsistencies. Furthermore, in a multi-camera system, the relative relationships between cameras need to be explicitly modeled to prevent pose optimization problems from becoming under-constrained.

% To address these challenges, we introduce a neural radiance field for under-calibrated multi-view cameras (UC-NeRF), a method designed to tackle the aforementioned challenges. 
To address these challenges, we introduce UC-NeRF, a method for high-quality neural rendering with multiple under-calibrated cameras.
We introduce three key innovations:
1) \textbf{Layer-based Color Correction}. To address color inconsistencies in the training images, especially for those taken by different cameras, we design a novel layer-based color correction module. This module separately adjusts the colors of the foreground and sky regions using two learned affine transformations for each image.  
2) \textbf{Virtual Warping}. We introduce a ``virtual warping" strategy that generates viewpoint-diverse yet color-consistent observations for each camera at each moment. These warped images under virtual viewpoints offer stronger constraints on the latent codes for color correction, especially for multi-camera systems where the overlapping region between cameras is limited. Moreover, the virtual warping strategy naturally expands the range of the training views for NeRF, enhancing its effectiveness in learning both the scene's appearance and geometry.
3) \textbf{Spatiotemporally Constrained Pose Refinement}. We propose a spatiotemporally constrained pose optimization strategy that explicitly models the spatial and temporal connections between cameras for pose optimization. This approach also improves the robustness against photometric differences by utilizing reprojection errors during pose optimization.

% Our approach highlights three novel modules,
% $i.e.$, layer-based color correction, virtual warping, and spatial-constrained pose refinement. First, to address the color discrepancies between different cameras, a layer-based color correction module adjusts the color of the foreground and sky separately with learned affine transformations for each image. An identity regularization term is also proposed to further constrain the color correction. Second, we propose a "virtual warping" strategy, which provides viewpoint-diverse but color-consistent observations for each camera at each moment. These observations provide stronger constraints on the latent codes for color correction. Moreover, virtual warping inherently broadens the range of training perspectives for NeRF, aiding NeRF in better learning both scene appearance and geometry. Third, we propose a spatial-constrained pose optimization strategy. We explicitly model spatial relationships between images to optimize poses. We enhance the robustness of photometric changes by using reprojection errors for pose optimization. 

Experiments on the public datasets Waymo (\cite{sun2020scalability} and NuScenes (\cite{caesar2020nuscenes}) show that our method achieves high-quality renderings with a multi-camera system and outperforms other baselines by a large margin.
Moreover, we show that the obtained high-quality renderings of novel views can facilitate downstream perception tasks like depth estimation.

% In experiments, we illustrate the prevalence of rendering degradation in a multi-camera setup across various NeRF algorithms. We demonstrate our method in two popular multi-camera datasets Waymo (~\cite{sun2020scalability} and NuScenes (~\cite{caesar2020nuscenes}), establishing the state-of-the-art of neural rendering in autonomous driving under multi-camera setups. We further employ the trained NeRF to render additional views and extend the training data for depth estimation. The results demonstrate an improvement in depth estimation performance, highlighting the high quality of our rendering images and the potential of novel view rendering in downstream perception tasks.

\vspace{-0.1cm}
\section{Related Work}
%\subsection{Multi-view Stereo}
\vspace{-0.1cm}
\paragraph{Multi-view Stereo}
Multi-view stereo (MVS) is a fundamental 3D vision task that aims to reconstruct a 3D model from posed images. Traditional methods (\cite{campbell2008using, furukawa2009accurate, bleyer2011patchmatch, furukawa2015multi, schonberger2016pixelwise}) exploit pixel correspondences between images from hand-crafted features to infer 3D structure. Deep learning methods (\cite{yao2018mvsnet, vakalopoulou2018atlasnet, long2020occlusion, chen2019point, long2021multi, ma2022multiview, feng2023cvrecon}) generally build multi-view correspondences implicitly and regress the 3D scenes as depth maps or 3D volumes in an end-to-end framework. 
Despite the increasing capability of MVS techniques in reconstructing accurate 3D models, it is strenuous to integrate 3D models into the traditional rendering pipeline to achieve photorealistic rendering. 

%\subsection{Neural Radiance Fields}
%General NeRF
% Neural Radiance Fields (NeRF), introduced by \cite{{mildenhall2021nerf}}, model the scene's volumetric structure and appearance as a continuous function, which enables accurate novel view synthesis in an efficient way. Subsequent works improve vanilla NeRF from different perspectives, such as scene parameterization strategy (~\cite{barron2022mip, wang2023f2}), acceleration (~\cite{muller2022instant, chen2022tensorf}), unbounded outdoor scenes (~\cite{tancik2022block, xie2022s}), and collaborative optimization of poses (~\cite{wang2021nerf, lin2021barf}), making it more effective in diverse application scenes. We review the most relevant works in the following part.
\vspace{-0.3cm}
\paragraph{NeRF for Outdoor Scenes}
NeRF (\cite{mildenhall2021nerf}) is a revolutionary technology that allows for the rendering of realistic images without the prerequisite of explicitly reconstructing 3D models.
It has demonstrated its effectiveness in high-quality novel view synthesis on indoor scenes and small-scale outdoor scenes. But it faces challenges when applied to unbounded outdoor scenes due to infinite depth range, complex illumination, and dynamic objects. To make NeRF more effective for infinite depth range, NeRF++ (\cite{zhang2020nerf++}) divides the scene space into foreground and background regions with an inverted sphere parameterization. 
In the following works (\cite{barron2022mip, wang2023f2}), more complicated non-linear scene parameterization is proposed to model the outdoor space more compactly and sample points more efficiently.
Some other works (\cite{deng2022depth, xie2022s, wang2023neural, yang2023unisim}) learn the complex geometry of outdoor scenes by introducing depth and surface normal priors. 
Moreover, to adapt to the view-dependent appearance due to surface reflection, camera parameters, and environment change, several works (\cite{martin2021nerf, rematas2022urban, tancik2022block, turki2022mega, li2023nerf}) learn appearance-related latent codes independently to control the view-dependent effect.
%NeRF in the wild. Urban Radiance field. Block-NeRF. 
Besides, some works (\cite{ xie2022s, turki2023suds}) model dynamic objects separately based on semantic priors, using 3D detection or semantic segmentation. 
In comparison, we primarily address the rendering quality deterioration problem caused by under-calibration of photometry and poses in a multi-camera setup for large-scale outdoor scenes.

%depth range: NeRF++, S-NeRF; Mip-NeRF360, F^2 NeRF; 
%lightning: NeRF in the wild, Urban Radiance filed; Block-NeRF
%dynamic NeRF:SUDS, Uni-Sim; Neural scenes graphs for dynamic scenes
% 
\vspace{-0.3cm}
\paragraph{NeRF with Pose Refinement}
% While taking the appearance of each point under multiple viewpoints as training data, NeRF heavily relies on accurate camera poses for the input images.
NeRF methods always require accurate camera poses to optimize a neural 3D scene.
However, the camera poses obtained from Structure-from-Motion (SfM) usually contain subtle errors that could significantly degrade the quality of the reconstructed NeRF.
NeRF$--$ (\cite{wang2021nerf}) jointly optimizes camera parameters with NeRF training via the photometric loss.
However, pose optimization struggles to achieve effective updates due to the increased non-linearity of NeRF arising from position encoding. BARF (\cite{lin2021barf}) eliminates this negative impact with a coarse-to-fine training strategy. SiNeRF (\cite{xia2022sinerf}) and GARF (\cite{shi2022garf}) replace the positional encoding with different activation functions to reduce non-linearity while maintaining the same rendering quality.
Besides, SCNeRF (\cite{jeong2021self}) and SPARF (\cite{truong2023sparf}) propose different geometric losses to improve the pose accuracy further. However, directly employing these methods in multi-camera systems will lead to a bottleneck, since the spatial relation between cameras is not taken into consideration. 
MC-NeRF (~\cite{gao2023mc}) is a contemporaneous work on multi-camera systems, but it focuses on optimizing the intrinsics of different cameras during pose optimization while we propose the spatiotemporal constraint between different cameras to enhance pose optimization.
% \lxx{don't mix pose and ISP, the two problems together.
% In the scenario of autonomous driving, under-calibrated multi-camera introduces additional errors related to inter-camera relative transformations and photometric inconsistencies stemming from ISP. 
% The former makes it challenging to achieve accurate optimization when independently modeling each camera's pose, while the latter violates photometric consistency during pose-NeRF joint optimization. Therefore, we propose a spatial-constrained pose optimization strategy that explicitly models the correlations among multiple cameras and leverages multi-view correspondences to refine poses.}
%summary the relationship between above works and our work
% Contrary to these methods that optimize poses based on photometric loss within the NeRF framework, we explicitly establish pixel correspondence to optimize poses considering color inconsistencies among multiple cameras in a multi-camera system.
\section{Method}

Our UC-NeRF extends the general NeRF algorithm to the multi-camera setup in autonomous driving. 
We begin by reviewing the common NeRF pipeline. Then we introduce the layer-based color correction (Sec.~\ref{sec:bc}) to reformulate the color rendering for handling the inconsistent color supervision in multi-camera systems.  In Sec.~\ref{sec:vw}, we introduce our virtual warping strategy to assist color correction by generating viewpoint-diverse but color-consistent images. Finally, the spatiotemporally constrained pose refinement is explained in Sec.~\ref{sec:pr}.

\dlt{
Our UC-NeRF extends the general NeRF algorithm to better adapt to the multi-camera setup in autonomous driving. 
We begin by reviewing the common NeRF pipeline. Then the layer-based color correction (Sec.~\ref{sec:bc}) is introduced to reformulate the color rendering for handling the inconsistent color supervision in multi-camera systems. A virtual warping strategy (Sec.~\ref{sec:vw}) is further proposed to assist color correction by generating viewpoint-diverse but color-consistent images. Finally, the spatiotemporally constrained pose refinement (Sec.~\ref{sec:pr}) is added to alleviate the pose errors in the multi-camera system.
}
% we introduce UC-NeRF with three main improvements: layer-based color correction, virtual warping, and spatial-temporal-constrained pose refinement. 
% With these modules, we make full use of multi-view photometric and geometric consistency to achieve state-of-the-art rendering in multi-camera setting.

\subsection{preliminary}
NeRF models a 3D scene as a continuous implicit function $\theta$ and regresses the density $\sigma$ and color $\mathbf{c} \in \mathbb{R}^{3}$ of every individual 3D point given its 3D coordinate $\mathbf{p} \in \mathbb{R}^{3}$ and a unit-norm viewing direction $\mathbf{d} \in \mathbb{R}^{2}$.
%which respectively indicate the presence or absence of scene content, and the visual appearance at that 3D point.
To synthesize a 2D image, NeRF employs volume rendering which samples a sequence of 3D points along a camera ray $\mathbf{r}$ as 
$\mathbf{I}(\mathbf{r})=\sum_{n=1}^N T_n \alpha_n \mathbf{c}_n$, where $T_n$
is the accumulated transmittance of the sampled points, $\alpha_n$ and $\mathbf{c}_n$ is the alpha value and the color of the sampled $n$-th point. Detailed definitions can be referred to NeRF (\cite{mildenhall2021nerf}).

% \quad T_n=\exp \left(-\sum_{m=1}^{n-1} \sigma_m (t_{m+1}-t_{m})\right),
% \begin{equation}
% \label{method:nerf}
% \mathbf{I}(\mathbf{r})=\sum_{n=1}^N T_n\left(1-e^{-\sigma_n (t_{n+1}-t_{n})}\right) \mathbf{c}_n, \quad T_n=\exp \left(-\sum_{m=1}^{n-1} \sigma_m (t_{m+1}-t_{m})\right),
% \end{equation}
% where $\mathbf{o}$ is the origin of the ray, $\mathbf{d}$ is the ray's direction, $t_n$ is the distance varying from the nearest to the farthest extents along the ray, $T_n$
% is the accumulated transmittance along the ray from the camera origin to $t_n$, $\mathbf{c}_n$ and $\sigma_n$ are the corresponding color and density of the sampled point in $t_n$.

To optimize NeRF, the photometric loss between the rendered color $\mathbf{I}(\mathbf{r})$ and the ground truth color $\hat{\mathbf{I}}(\mathbf{r})$ from a set of sampled rays $\mathcal{R}$ is applied as:
\begin{equation}
\label{eq:phloss}
\mathcal{L}_{\text {pho}}(\theta)=\sum_{\mathbf{r} \in \mathcal{R}}\left\|\hat{\mathbf{I}}(\mathbf{r})-\mathbf{I}\left(\mathbf{r}\right)\right\|_2^2.
\end{equation}
%where $\mathcal{R}$ indicates a batch of rays for training.

% \subsection{Neural rendering from Multi-cameras}
% For autonomous driving, the cars are equipped with multiple cameras, which also benefits the training of NeRF by providing more views. However, the simple combination of images from different views can be problematic, as shown in Fig.~\ref{}.

\subsection{Layer-based Color Correction}
\label{sec:bc}

\begin{figure*}[t!]
\centering    %
\includegraphics[width=\linewidth]{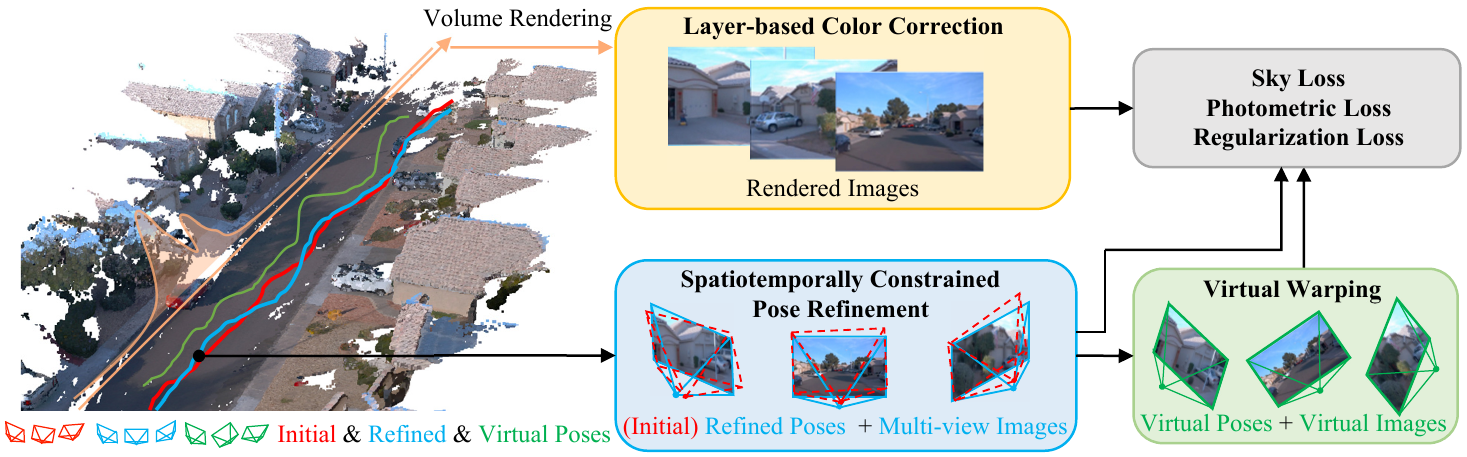}
\caption{Overview of UC-NeRF framework. To mitigate the inconsistency of color supervision in multi-camera systems, the spatiotemporally constrained pose refinement module optimizes poses and the layer-based color correction module models the image-dependent appearance from varying cameras and timestamps. The virtual warping module generates diverse virtual views with geometric and color consistency, enriching data for color correction and 3D scene recovery.} %
\label{fig: frame1}
\end{figure*}

% The imaging process of ISP units is quite complex, which involves various non-linear operations, such as auto-exposure, color space conversion, gamma correction, and so on.
In multi-camera systems, different cameras always have distinct Image Signal Processor (ISP) configurations, resulting in inconsistent imaging colors for the same 3D region. As a result, optimizing a NeRF representation using such inconsistent images always causes low-quality renderings.
%ISP unit is a sophisticated, non-linear system that converts raw data captured from visible light into viewable RGB images through a series of complex operations.
The work Urban-NeRF (\cite{ rematas2022urban}) attempts to approximate a global linear compensation transformation for each view to alleviate the discrepancies of the views from different cameras. 
% the discrepancies of different ISP units via optimizing a global linear transformation for each view.
It's worth noticing that due to the non-linear property of the ISP process, the pixels with different intensities in a single image even have various ISP imaging effects, so it is insufficient to model these spatially varying color patterns using a single global compensation transformation.
To balance quality and efficiency, we propose to split the scene into foreground-sky layers and model the color compensation transformation for each layer separately. 
This is because the sky regions are always much brighter than the foreground objects, and they present distinct ISP imaging effects.

We first model the foreground and sky as two independent NeRF models $\theta_{fg}$ and $\theta_{sky}$. The color of a rendered pixel from ray $\mathbf{r}$ is obtained by the weighted combination of foreground color $\mathbf{I}_{fg}(\mathbf{r})$ and sky color $\mathbf{I}_{sky}(\mathbf{r})$, as illustrated in Eq.~\ref{method:nerfrevise1}:
\begin{equation}
\label{method:nerfrevise1}
\mathbf{I}(\mathbf{r}) = \mathbf{I}_{fg}(\mathbf{r}) + (1-o_{fg}) \mathbf{I}_{sky}(\mathbf{r}),
\end{equation}
where $o_{fg} = \sum_{n=1}^N T_{n,fg}\alpha_{n,fg}$ is the accumulated weight of foreground NeRF in $\mathbf{r}$.
To encourage $o_{fg}$ approaches $1$ in the foreground area while approaches $0$ in the sky area, a binary cross-entropy loss is employed as Eq.~\ref{method:skyloss}:
\begin{equation}
\label{method:skyloss}
L_{sky}(o_{fg}, m_{sky})=-m_{sky} \log (1-o_{fg})-(1-m_{sky}) \log (o_{fg}),
\end{equation}
where $m_{sky}$ is the sky mask generated from pretrained segmentation model (\cite{yin2022devil}).

After modeling the foreground and sky NeRF, we approximate the color correction of the foreground and the sky using separate affine transformations.
Considering the color variance across both cameras and timestamps,
for each training image $\mathbf{I}_{i,k}$ from camera $k$ at timestamp $i$, 
% $\boldsymbol{\mu}_{i,k}$ $\boldsymbol{\lambda}_{i,k}$
a foreground correction code and a sky correction code are assigned to represent the image-dependent color variation (\textit{Subscripts $i, k$ are omitted in the following descriptions for clarity}). 
These correction codes are further decoded by a multi-layer perceptron (MLP) as the affine transformations $[\mathbf{A}, \mathbf{x}]$ and $[\mathbf{C}, \mathbf{y}]$, where $\mathbf{A}, \mathbf{C} \in \mathbb{R}^{3 \times 3}$, and $\mathbf{x}, \mathbf{y} \in \mathbb{R}^{3 \times 1}$. 
For the rendered pixel which emits $\mathbf{r}$ in $\mathbf{I}$, the final pixel color in Eq.~\ref{method:nerfrevise1} can be rewrited as:
\begin{equation}
\label{method:nerfrevise2}
\mathbf{I}(\mathbf{r}) = \mathbf{A} \mathbf{I}_{fg}(\mathbf{r}) + \mathbf{x} + (1-o_{fg})( \mathbf{C} \mathbf{I}_{sky}(\mathbf{r}) +  \mathbf{y}).
\end{equation}
%
% However, without any constraints on $\phi(\boldsymbol{\mu}), \phi(\boldsymbol{\lambda})$, the result %can 
% may deviate from the true color. %, as shown in Fig.~\ref{fig: l1reg}. 
% Fig.~\ref{fig: l1reg} (b) shows the rendered car's color deviates from the ground truth (Fig.~\ref{fig: l1reg} (a)) noticeably. 
%The car in (b) indicated by the red arrow exhibits a noticeable color deviation compared to the ground truth image (a).
% find a theory to introduce l1 regularization
To stabilize the optimization process and ensure that the adjusted color does not significantly deviate from the origin, we add a regularization term, as illustrated in Eq.~\ref{method:l1reg}:
\begin{equation}
\label{method:l1reg}
L_{reg}=\left|\mathbf{A}-\mathbf{E}_3\right| + \left|\mathbf{C}-\mathbf{E}_3\right| + \left|\mathbf{x}\right| + \left|\mathbf{y}\right|,
\end{equation}
where $\mathbf{E}_3$ refers to the identity matrix. 
% As shown in Fig.~\ref{fig: l1reg} (c), the color of the car matches that of the ground truth after identity regularization. Intuitively, the regularization term encourages minor adjustments to the color correction for image color. 

% \begin{figure*}[t!]
% \centering    %
% \includegraphics[width=\linewidth]{figures/l1reg.pdf}
% \caption{Color correction with or without regularization.
% The result with regularization (b) achieves more accurate colors than that without regularization (a).
% % the corrected color of the car (b) may deviate from its true color (a). This issue can be resolved by introducing an identity regularization term.
% } %
% \label{fig: l1reg}
% \end{figure*}

\subsection{Virtual Warping} 
\label{sec:vw}
\begin{figure*}[t!]
\centering    
\includegraphics[width=\linewidth]{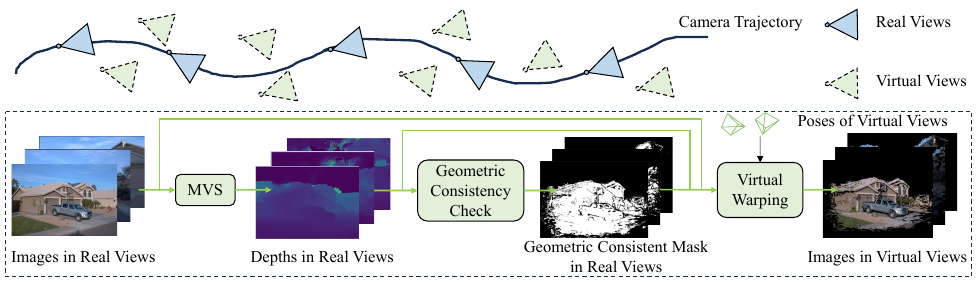}
\caption{
The generation of virtual warped images.
% Virtual warping generates diverse virtual views with excellent geometric and color consistency. 
For each known viewpoint, we generate its depth map using MVS and filter out inaccurate depths through a geometric consistency check. 
Each virtual view is obtained by warping the image from a known viewpoint to the virtual viewpoint.}
% that perturbed from the known viewpoint.
\label{fig: vw}
\end{figure*}

In multi-camera systems, images from different viewpoints often have limited overlapping areas, making it more challenging to align their colors compared to aligning frames from a single camera.
To align the image colors of multiple cameras and prevent the optimized latent codes for color correction from overfitting to a specific viewpoint, we propose virtual warping, which simulates more diverse yet color-consistent images under a set of virtual viewpoints for training. Furthermore, virtual warping naturally expands the range of perspectives available to NeRF, thereby enhancing its capability to reconstruct the 3D scene. Fig.~\ref{fig: vw} shows the pipeline of our virtual warping strategy. 
We employ an MVS method (\cite{ma2022multiview}) to generate depth maps of all views. To remove outliers and retain the consistent depths across multiple views, we further leverage a geometric consistent check process (\cite{schonberger2016pixelwise}) to generate a mask $\mathbf{M}$ that only keeps reliable depth values in each view. 

%
%As illustrated in Fig.~\ref{fig: vw}, to create virtual views, we initially estimate the depth map for each training image using MVS. Then we employ geometric consistency filtering to retain depth values that remain consistent across multiple views. Finally, we perform pixel warping from observed views to virtual views to generate pseudo-RGB data for training NeRF.
% Besides, our virtual warping also helps the NeRF better render novel views that deviate from the original camera trajectory.
%
% Specifically, for a pixel $\mathbf{p}_{r}$ in a reference view with estimated depth $d_{r}$, the depth $\mathbf{d}_{r \rightarrow i}$ of the warped pixel $\mathbf{p}_{r \rightarrow i}$ from reference view to a target view $i$ can be induced by: 

% where $\overline{\mathbf{p}}_{r \rightarrow i}$ and $\overline{\mathbf{p}}_{r}$ is the homogeneous coordinates of $\mathbf{p}_{r \rightarrow i}$ and $\mathbf{p}_{r}$, $\mathbf{K}$ is the camera intrinsic, $[\mathbf{R}_{r \rightarrow i}, \mathbf{t}_{r \rightarrow i}]$ is the relative transformation from reference view to target view $i$.
% For each reference view, we choose $m$ target views that are near the reference view and generate a geometric consistency mask $\mathbf{M}$ as:
% \begin{equation}
% \mathbf{M}(\mathbf{p}_r)= \left\{\begin{array}{l}
% 0, \text { if } \left(\sum_{i=0}^m \beta\left(\left|d_{r \rightarrow i}-d_i\right| / d_i\right)\right)<\gamma \\
% 1, \text { otherwise. }
% \end{array}\right.
% % 
% , \beta(x)=\left\{\begin{array}{l}
% 0, \text { if } x \geq \alpha \\
% 1, \text { otherwise. }
% \end{array}\right. ,
% \end{equation}
% where pixel is retained if it appears in at least $\gamma$ target views, and the absolute relative error between its warped depth and depth in target view is within $\alpha$.

%\paragraph{Warping with Virtual Poses}
%Upon obtaining depth information and establishing a consistent mask, we proceed to generate virtual poses by introducing controlled perturbations to the original trajectory. 
With estimated depths, we generate multiple virtual poses and warp colors and color correction codes to the virtual positions. Specifically, we perturb an existing pose $\mathbf{T}_{o}$ with an additional transformation $[\mathbf{R}_{o \rightarrow v}, \mathbf{t}_{o \rightarrow v}]$ as a virtual pose $\mathbf{T}_{v}$.
% with $\Delta \mathbf{P}$, i.e. $\mathbf{T}_{v}=\mathbf{T}_{o}\Delta \mathbf{P}$. 
The rotation $\mathbf{R}_{o \rightarrow v}$ is generated by randomly selecting one of the three axes with a random angle $\in [-20^{\circ}, 20^{\circ}]$. The translation $\mathbf{t}_{o \rightarrow v}$ is a 3D vector of random direction with a length  $\in [0m,1m]$. 
%
%For each existing pose $\mathbf{T}_{o}$ along the trajectory, a virtual pose $\mathbf{T}_{v}$ is created by multiply the additional transformations as $\mathbf{T}_{v}=\mathbf{T}_{o}\Delta \mathbf{P}$.
% where $\Delta \mathbf{P}$ contains a randomly sampled rotation within the range of $[5, 20]$ degree and translation within the range of $[0, 1]$ meter. The directions of rotation and translation are randomly selected from the positive and negative directions of the three coordinate axes in the camera coordinate system. We randomly sample $v$ virtual poses for each existing pose.
%For each virtual pose, we warp the pixel $\mathbf{p}$ in its original pose to the virtual pose, which is the same as the warping process in Eq.~\ref{method:warping}.
% \cxj{How do you sample the virtual poses? What is the range of translation and rotation? How many virtual poses are generated? Each original pose (Cik) will produce a number of virtual poses and all the new images are generated according to this single original pose? Or a single virtual image is synthesized from a number of images from multiple original poses? Fig. 4 is a little bit consufing.  }
%write this two part in the math form.
% Considering object occlusions, some different pixels in the original view will be warped to the same pixel in the virtual view, where we preserve the warped pixel with minimum depth.
Each pixel $\mathbf{p}_{o}$ in an existing image taken under camera pose $\mathbf{T}_{o}$ is warped to an image point $\mathbf{p}_{v}$ with the virtual pose $\mathbf{T}_{v}$ as:
\begin{equation}
\label{method:warping}
d_{v} \overline{\mathbf{p}}_{v} = \mathbf{K}(\mathbf{R}_{o \rightarrow v} \mathbf{K}^{-1} d_{o} \overline{\mathbf{p}}_{o}+\mathbf{t}_{o \rightarrow v}) ,
\end{equation}
where $\overline{\mathbf{p}}_{o}$ and $\overline{\mathbf{p}}_{v}$ is the homogeneous coordinates of $\mathbf{p}_{o}$ and $\mathbf{p}_{v}$, $\mathbf{K}$ is the camera intrinsic matrix, $d_v$ and $d_o$ are the pixel depth in the virtual view and the corresponding pixel depth in the original real view.
Considering object occlusions, there could be multiple pixels in the original views mapped to the same position in the virtual view, so we keep the warped pixel with the minimum depth. 

After warping, the geometric consistency mask $\mathbf{M}$ is applied to the warped pixels to filter pixels with noisy depth.
Then the color and the color correction code of the pixel in the original view are assigned to the corresponding warped pixels in the virtual view. This provides more clues to recover the consistent appearance and geometry of the scene.

\begin{wrapfigure}{r}{0.3\textwidth}
\vspace{-0.2cm}
\includegraphics[width=1\linewidth]{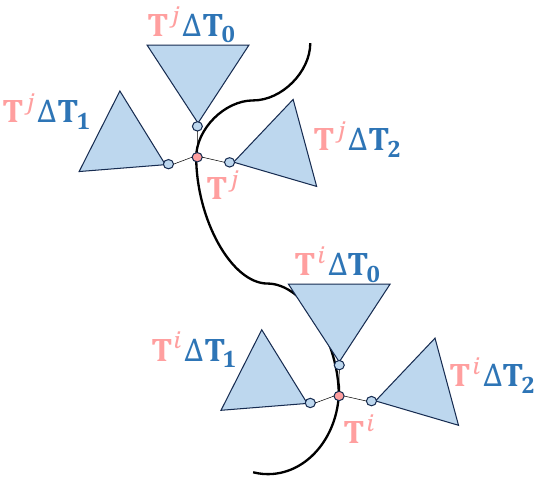}
\caption{Pose modeling.} 
\label{fig:cameras}
\vspace{-0.4cm}
\end{wrapfigure}

\subsection{Spatiotemporally constrained Pose Refinement}

\label{sec:pr}
The rendering quality of NeRF heavily relies on the accuracy of camera poses. 
Previous approaches (\cite{tancik2022block, xie2022s}) model the camera poses independently and jointly optimize them within the NeRF framework. They do not fully exploit the spatial correlations between cameras in multi-camera systems, leading to under-constrained pose optimization.
Additionally, the camera pose optimization depends on the photometric consistency assumption, which is usually violated in long-time videos captured in multi-camera systems. 
Given the condition that the cameras have a fixed spatial relationship with the main capturing device (i.e. the driving car) during the whole process, we explicitly establish the temporally fixed geometric transformation between cameras. 

While capturing multi-view images with $K$ cameras, the pose $\mathbf{T}_{k}^{i}$ of the $k$th camera at time $i$ is denoted as the combination of the car's ego pose $\mathbf{T}^{i}$ and the relative transformation $\Delta \mathbf{T}_{k}$, which is temporally consistent and optimizable, as Fig.~\ref{fig:cameras} shows. 
%The transformations $\{\Delta \mathbf{T}_{k}\}_{k=1,\ldots,K}$ model the \textbf{spatial} relationship between cameras and are \textbf{temporally} consistent. 
%The advantage of spatial relationship
Explicitly modeling the spatial relationship between cameras provides more restrictions for pose refinement, thus effectively enhancing robustness against incorrect point matches across all frames.

After building the spatio-temporal constraint between camera poses, point correspondences between images captured by different cameras at different times are established as correlation graph $\mathcal{E}$. 
Then we employ bundle adjustment to minimize the reprojection error defined as:
\begin{equation}
\label{eq:ba}
L_{rpj}=\sum_{((i,k), (j,l)) \in \mathcal{E}}\left\| \mathbf{p}_{l}^{j}-\Pi_{l}\left((\mathbf{T}^{j} \Delta \mathbf{T}_{l})^{-1} \mathbf{T}^{i} \Delta \mathbf{T}_{k} \Pi_{k}^{-1}\left(\mathbf{q}^{i}_{k}\right)\right)\right\|^2,
\end{equation}
where $\mathbf{p}^{j}_{l}$ and $\mathbf{q}^{i}_{k}$ are pixels in the images captured by camera $l$ at time $j$ and camera $k$ at time $i$, $\Pi_{l}$ and $\Pi_{k}^{-1}$ are projection function of camera $l$ and unprojection function of camera $k$.

\subsection{Training Strategy}
\label{sec:training}

Our training strategy mainly includes two parts: 1) Pose refinement and depth estimation.
We initialize the poses from sensor-fusion SLAM and further optimize them using our proposed spatiotemporally constrained pose refinement module, as described in Eq.~\ref{eq:ba}. With these refined poses, we generate a depth map and geometric consistency mask for each image, following the procedure outlined in Sec.~\ref{sec:vw}. 2) End-to-end NeRF optimization. Specifically, the proposed layer-based color correction and virtual warping are used in the optimization of NeRF to achieve high-quality renderings. 
In each training batch, we randomly sample $B$ real images and employ our virtual warping module to create $V$ virtual views for each real image. The pixels are randomly sampled from these real and virtual views as the ground truth for NeRF training. Our UCNeRF renders these pixels based on Eq.~\ref{method:nerfrevise2} and is supervised by the loss function in Eq.~\ref{totalloss}:

\begin{equation}
\label{totalloss}
L=L_{p h o}+\lambda L_{\text {sky }}+\gamma L_{\text {reg}},
\end{equation}

where $\lambda$ and $\gamma$ are the weights of $L_{\text {sky }}$ and $L_{\text {reg}}$.
\section{Experiments}

%主表: 在waymo和Nuscenes上多个camera渲染结果
\begin{table*}
  \centering
  \footnotesize
  \caption{Comparison on Waymo and NuScenes. Our method significantly outperforms state-of-the-art methods in both datasets and all the evaluation metrics.}
  \vspace{0.1cm}
    \newcolumntype{"}{@{\hskip\tabcolsep\vrule width 1.2pt\hskip\tabcolsep}}
  \label{tab:urbancomparison}
  \begin{tabular}{@{}l|ccc"ccc@{}}
    \toprule
    & & Waymo & & & NuScenes & \\
    %\hline
    Method & PSNR $\uparrow$ & SSIM $\uparrow$ & LPIPS $\downarrow$ & PSNR $\uparrow$ & SSIM $\uparrow$ & LPIPS $\downarrow$ \\
  \hline
    Mip-NeRF (\cite{barron2021mip}) & 22.42 & 0.698 & 0.471 & 23.31 & 0.758 & 0.489 \\
    Mip-NeRF 360 (\cite{barron2022mip}) & 24.46 & 0.769 & 0.406 & 25.15 & 0.809 & 0.436 \\
    % Urban-NeRF (~\cite{}) & & & & & & \\
    Instant-NGP (\cite{muller2022instant}) & 23.84 & 0.702 & 0.494 & 23.81 & 0.777 & 0.476 \\
    S-NeRF (\cite{xie2022s}) & 24.89 & 0.772 & 0.401 & 26.02 & 0.824 & 0.415  \\
    % 3D Gaussian Splatting (~\cite{kerbl20233d}) & 25.01 & 0.794 & 0.368 & 22.69 & 0.772 & 0.501 \\
    Zip-NeRF (\cite{barron2023zip}) & 26.21 & 0.815 & 0.389 & 27.06 & 0.831 & 0.435 \\
    UC-NeRF (Ours) & \textbf{28.13} & \textbf{0.842} & \textbf{0.356} & \textbf{30.20} & \textbf{0.876} & \textbf{0.374} \\
    \bottomrule
  \end{tabular}
\end{table*}

\subsection{Datasets and Implementation Details}
\label{exp:dataset}
\paragraph{Datasets.} We conduct our experiments in two urban datasets which contain images captured from multi-cameras, $i.e.$, Waymo (\cite{sun2020scalability}) and NuScenes (\cite{caesar2020nuscenes}). 
% The only prior work utilized the multi-camera setup in autonomous driving, S-NeRF (~\cite{xie2022s}), models dynamic objects and is evaluated in selected dynamic scenes. However, our focus in this work is primarily on multi-camera aspects, without considering dynamic objects.
% We conduct our experiments on two urban datasets that contain images captured from multi-cameras, $i.e.$, Waymo (~\cite{sun2020scalability}) and NuScenes (~\cite{caesar2020nuscenes}). 
We select ten static scenes in Waymo and five static scenes in NuScenes for evaluation. 
To evaluate the performance of novel view synthesis, following common settings, we select one of every eight images of each camera as testing images and the remaining ones as training data. 
We apply the three widely-used metrics for evaluation, $i.e.$, peak signal-to-noise ratio (PSNR), structural similarity index measure (SSIM), and the learned perceptual image patch similarity (LPIPS) (\cite{zhang2018unreasonable}).

\paragraph{Baselines.} We choose Zip-NeRF (\cite{barron2023zip}) as our baseline. Since there is no official code for Zip-NeRF, we use the reimplementation of Zip-NeRF in \cite{zipcode}. We compare our method with the baseline and other state-of-the-art NeRF methods, including Mip-NeRF (\cite{barron2021mip}), Mip-NeRF 360 (~\cite{barron2022mip}), Instant-NGP (\cite{muller2022instant}), and S-NeRF (\cite{xie2022s}). We provide implementation details in the appendix.

% two popular NeRF algorithms: Zip-NeRF (~\cite{barron2023zip}) and 3D Gaussian Splatting (~\cite{kerbl20233d}), illustrating the ubiquity of the deterioration for NeRF's rendering with the incorporation of multiple cameras and demonstrating the versatility of our approach across various neural rendering representations.
% %rendering loss; the detail of virtual warp;
% Please refer to supplementary materials for more details.

\subsection{Results on novel view synthesis}

\begin{figure*}
\vspace{0.2cm}
\centering    
\includegraphics[width=\linewidth]{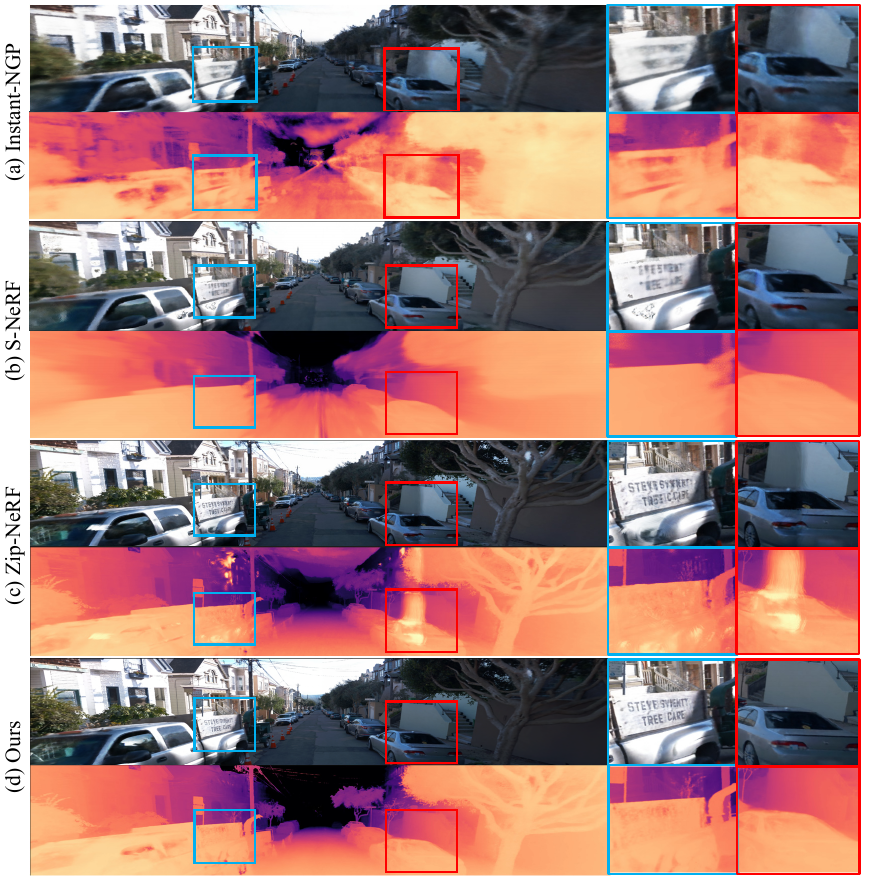}
\caption{We render the panoramas at a resolution of $5760\times 1280$ for comparison. Notable enhancements are indicated in blue and red boxes, and cropped patches are displayed to emphasize specific details. Compared to other methods, our results present consistent color and sharp details, even faithfully recovering the slogans. For additional results, please refer to the appendix.} 
\label{fig: maincompare}
% \vspace{-0.2cm}
\end{figure*}

The comparison result of neural rendering in urban scenes with a multi-camera setting is shown in Tab.~\ref{tab:urbancomparison}. 
With the proposed layer-based color correction, virtual warping, and spatio-temporally constrained pose refinement, our UC-NeRF outperforms the other methods in both datasets. We also show the panoramic rendering results in Fig.~\ref{fig: maincompare}. 
Without any anti-aliasing design in the rendering process, images generated by Instant-NGP and S-NeRF exhibit notable blurriness. 
Although Zip-NeRF features an anti-aliasing mechanism, it also amplifies the artifacts caused by inconsistent color supervision across different views.  
Our approach excels at rendering consistent colors and sharp details, as highlighted in the regions of texts, cars, and buildings.  
Additionally, our method provides more accurate 3D reconstruction, as demonstrated by the depth maps. 
We show more results on Waymo and NuScenes in the appendix.

\subsection{Ablation study}

\begin{wraptable}{r}{0.5\textwidth}
\vspace{-1.8cm}
\centering
\footnotesize
\caption{Ablation study.}
\label{tab:mainablation}
\begin{tabular}{@{}ccc|lll@{}}
\toprule
LCC & \multicolumn{1}{c}{STPR} & VW & PSNR $\uparrow$ & SSIM $\uparrow$ & LPIPS $\downarrow$ \\ \midrule
 \XSolidBrush  & \XSolidBrush &  \XSolidBrush  
 & 26.21 & 0.815 & 0.389 \\
 \XSolidBrush  &  \Checkmark &  \XSolidBrush  
 & 26.95 & 0.839 & 0.360  \\
 \Checkmark & \XSolidBrush & \XSolidBrush  
 &  27.18 &  0.820  &  0.375    \\
  \Checkmark  &  \XSolidBrush  & \Checkmark 
  & 27.26 & 0.825 & 0.372 \\
  \Checkmark  &  \Checkmark  & \XSolidBrush 
  & 27.82 & 0.838 & 0.371 \\
 \Checkmark  & \Checkmark & \Checkmark 
 &  \textbf{28.13} & \textbf{0.842} & \textbf{0.356} \\ \bottomrule
\end{tabular}

\end{wraptable}

\begin{figure*}
\vspace{-0.4cm}
\centering    
\includegraphics[width=\linewidth]{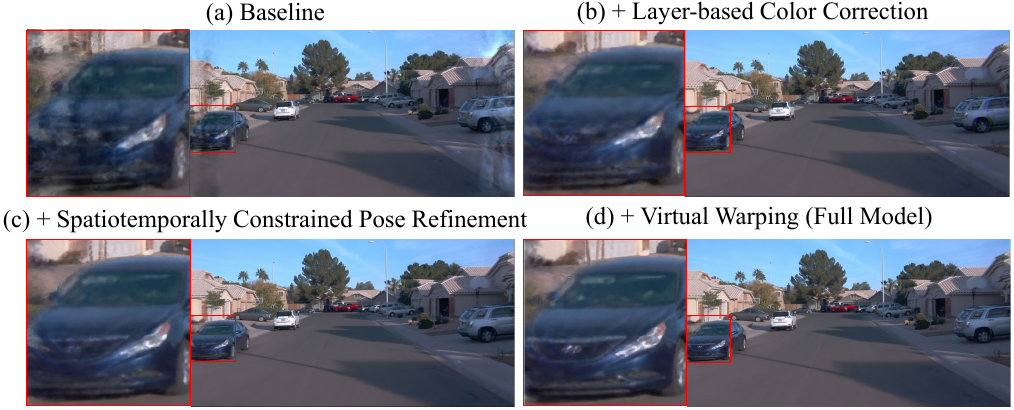}
\caption{Each proposed module progressively enhances the rendering quality of novel views.} 
\label{fig: mainablation}
\end{figure*}

We conduct extensive ablation studies on ten scenes from the Waymo dataset to explore the effect of each proposed module in our UC-NeRF. 
We investigate the effect of each module, $i.e.$, layer-based color correction (LCC), spatiotemporally constrained pose refinement (STPR), and virtual warping (VW).
As shown in Tab.~\ref{tab:mainablation}, the layer-based color correction module brings significant improvement (the third row) compared with the baseline model, since it solves the problem of inconsistent color supervision between views in training. 
Fig.~\ref{fig: mainablation} (b) also illustrates that the LCC module reduces hazy artifacts and presents sharper renderings.
By incorporating the spatiotemporally constrained pose refinement (STPR) module, the quality of rendering is further improved.
% the significant reduction of hazy artifacts compared to the baseline model (a) with the incorporation of LCC. By further implementing spatial-temporal-constrained pose refinement (SPR) to reduce pose errors, we achieve an additional enhancement in rendering quality (the second, fifth row in Tab.~\ref{tab:mainablation}). Compared (c) with (b) in Fig.~\ref{fig: mainablation}, SPR effectively eliminates the ghosting artifacts in the rendered images. 
Moreover, our virtual warping (VW) strategy can enrich the diversity of training views for learning color correction, appearance, and geometry. Even the object details, e.g. the car lights and the car emblem, become more discernible in  Fig.~\ref{fig: mainablation} (d). One notable thing is that the accuracy of the virtual views provided by virtual warping is closely related to the accuracy of the poses. Thus, virtual warping provides a more noticeable boost when the pose refinement module is added (the fourth and sixth row in Tab.~\ref{tab:mainablation}).
More detailed discussions can be referred to in the appendix.
% ablation studies about the layer-based color correction and temporal-spatial-constrained pose refinement with existing methods can be referred to in the appendix.

\begin{figure*}
\vspace{-0.2cm}
\centering    
\includegraphics[width=\linewidth]{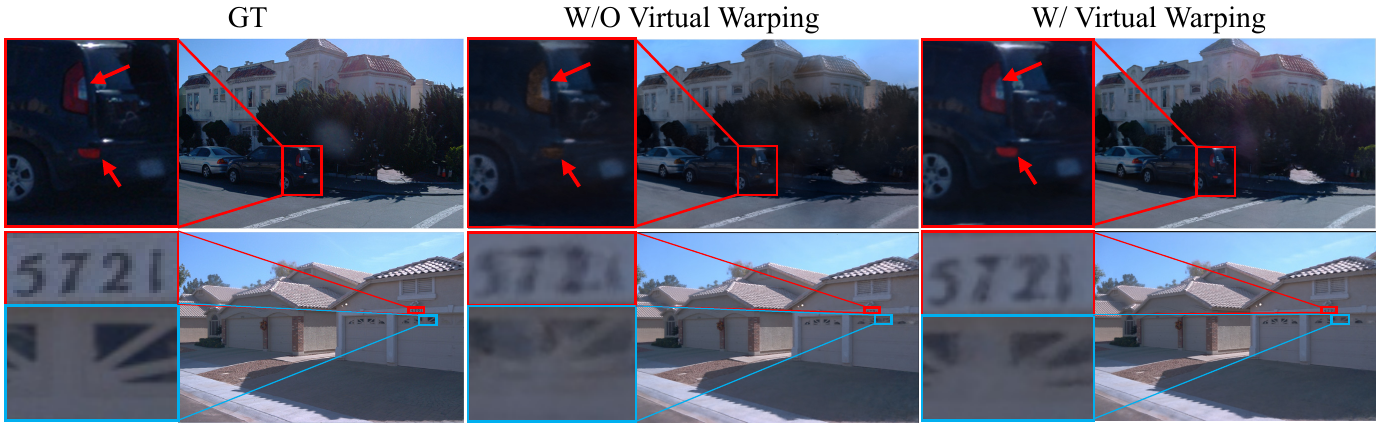}
\caption{Virtual warping benefits color correction (top) and edge sharpness (bottom).} 
\vspace{-0.3cm}
\label{fig: vwablation}
\end{figure*}

\paragraph{Benifits of Virtual Warping}
Virtual warping enriches NeRF's training perspectives of each camera by generating images with consistent geometry and photometry. In addition to the overall improvement in rendering quality shown in Tab.~\ref{tab:mainablation}, we present more cases demonstrating a significant enhancement in rendering quality after incorporating virtual warping in Fig.~\ref{fig: vwablation}. This includes better color correction (the first row) and enhanced image details (the second row).

\begin{wraptable}{r}{0.65\textwidth}
\label{efficiency}
\centering
\footnotesize
\vspace{-0.4cm}
\caption{Efficiency Analysis. Tested on one NVIDIA Tesla V100 GPU with image resolution $1920 \times 1280$.}
\begin{tabular}{@{}l|cccc@{}}
\toprule
Method & Training & Inference & PSNR\\ 
\midrule
Mip-NeRF (\cite{barron2021mip}) & 20h & 70s & 22.42 \\
Mip-NeRF-360 (\cite{barron2022mip})  & 14h & 42s & 24.46 \\
Instant-NGP (\cite{muller2022instant}) & 30min & 0.35s  & 23.84 \\
S-NeRF (\cite{xie2022s}) & 15h & 80s &  24.89 \\
Zip-NeRF (\cite{barron2023zip}) & 2h & 2s & 26.21 \\
UC-NeRF (Ours) & 3h & 3.2s & 28.13 \\ \bottomrule
\end{tabular}
\vspace{-0.3cm}
\label{tab:efficiency}
\end{wraptable}

\paragraph{Efficiency Analysis}

We compare the efficiency of different methods in Tab.~\ref{tab:efficiency}. All methods are tested on one NVIDIA Tesla V100 GPU with an image resolution of $1920 \times 1280$. Note that our training time includes both steps (pose refinement and NeRF training) described in Sec.~\ref{sec:training}.
Zip-NeRF is more efficient than other methods except Instant-NGP, which is specifically designed for NeRF acceleration. 
Since our method is built upon Zip-NeRF, our method consumes a bit more time than Zip-NeRF but achieves a significant improvement in rendering quality.

\section{Conclusion and Future Work}
In conclusion, we propose UC-NeRF that effectively addresses the challenges of integrating multi-camera systems into the NeRF paradigm. 
Experiments on Waymo and NuScenes demonstrate a significant improvement in rendering quality, setting a new benchmark for neural rendering within multi-camera setups. Furthermore, the application of our trained NeRF for improving depth estimation has shown promising results, underscoring the high rendering quality of novel views and the potential of NeRF for downstream perception tasks.
% Future work may focus on extending our framework in real-time rendering and exploring its applicability to other perception tasks. 
Looking forward, there are several promising avenues for future exploration. One key objective is to adapt our method to a real-time rendering framework.
This advancement would be of great value in autonomous driving, where real-time simulation and interaction are critical.
Besides, the promising results obtained from depth estimation invite further exploration into the versatility of our NeRF model. It would be worthwhile to investigate if the high-quality rendering can be utilized to enrich other perception tasks, e.g. semantic segmentation and object detection, thereby broadening the scope of our model's applications.

\bibliography{conference}

\begin{thebibliography}{49}
\providecommand{\natexlab}[1]{#1}
\providecommand{\url}[1]{\texttt{#1}}
\expandafter\ifx\csname urlstyle\endcsname\relax
  \providecommand{\doi}[1]{doi: #1}\else
  \providecommand{\doi}{doi: \begingroup \urlstyle{rm}\Url}\fi

\bibitem[Barron et~al.(2021)Barron, Mildenhall, Tancik, Hedman, Martin-Brualla, and Srinivasan]{barron2021mip}
Jonathan~T Barron, Ben Mildenhall, Matthew Tancik, Peter Hedman, Ricardo Martin-Brualla, and Pratul~P Srinivasan.
\newblock Mip-nerf: A multiscale representation for anti-aliasing neural radiance fields.
\newblock In \emph{Proceedings of the IEEE/CVF International Conference on Computer Vision}, pp.\  5855--5864, 2021.

\bibitem[Barron et~al.(2022)Barron, Mildenhall, Verbin, Srinivasan, and Hedman]{barron2022mip}
Jonathan~T Barron, Ben Mildenhall, Dor Verbin, Pratul~P Srinivasan, and Peter Hedman.
\newblock Mip-nerf 360: Unbounded anti-aliased neural radiance fields.
\newblock In \emph{Proceedings of the IEEE/CVF Conference on Computer Vision and Pattern Recognition}, pp.\  5470--5479, 2022.

\bibitem[Barron et~al.(2023)Barron, Mildenhall, Verbin, Srinivasan, and Hedman]{barron2023zip}
Jonathan~T Barron, Ben Mildenhall, Dor Verbin, Pratul~P Srinivasan, and Peter Hedman.
\newblock Zip-nerf: Anti-aliased grid-based neural radiance fields.
\newblock \emph{Proceedings of the IEEE/CVF International Conference on Computer Vision}, 2023.

\bibitem[Bleyer et~al.(2011)Bleyer, Rhemann, and Rother]{bleyer2011patchmatch}
Michael Bleyer, Christoph Rhemann, and Carsten Rother.
\newblock Patchmatch stereo-stereo matching with slanted support windows.
\newblock In \emph{Bmvc}, volume~11, pp.\  1--11, 2011.

\bibitem[Caesar et~al.(2020)Caesar, Bankiti, Lang, Vora, Liong, Xu, Krishnan, Pan, Baldan, and Beijbom]{caesar2020nuscenes}
Holger Caesar, Varun Bankiti, Alex~H Lang, Sourabh Vora, Venice~Erin Liong, Qiang Xu, Anush Krishnan, Yu~Pan, Giancarlo Baldan, and Oscar Beijbom.
\newblock nuscenes: A multimodal dataset for autonomous driving.
\newblock In \emph{Proceedings of the IEEE/CVF conference on computer vision and pattern recognition}, pp.\  11621--11631, 2020.

\bibitem[Campbell et~al.(2008)Campbell, Vogiatzis, Hern{\'a}ndez, and Cipolla]{campbell2008using}
Neill~DF Campbell, George Vogiatzis, Carlos Hern{\'a}ndez, and Roberto Cipolla.
\newblock Using multiple hypotheses to improve depth-maps for multi-view stereo.
\newblock In \emph{Computer Vision--ECCV 2008: 10th European Conference on Computer Vision, Marseille, France, October 12-18, 2008, Proceedings, Part I 10}, pp.\  766--779. Springer, 2008.

\bibitem[Chen et~al.(2019)Chen, Han, Xu, and Su]{chen2019point}
Rui Chen, Songfang Han, Jing Xu, and Hao Su.
\newblock Point-based multi-view stereo network.
\newblock In \emph{Proceedings of the IEEE/CVF international conference on computer vision}, pp.\  1538--1547, 2019.

\bibitem[Deng et~al.(2022)Deng, Liu, Zhu, and Ramanan]{deng2022depth}
Kangle Deng, Andrew Liu, Jun-Yan Zhu, and Deva Ramanan.
\newblock Depth-supervised nerf: Fewer views and faster training for free.
\newblock In \emph{Proceedings of the IEEE/CVF Conference on Computer Vision and Pattern Recognition}, pp.\  12882--12891, 2022.

\bibitem[DeTone et~al.(2018)DeTone, Malisiewicz, and Rabinovich]{detone2018superpoint}
Daniel DeTone, Tomasz Malisiewicz, and Andrew Rabinovich.
\newblock Superpoint: Self-supervised interest point detection and description.
\newblock In \emph{Proceedings of the IEEE conference on computer vision and pattern recognition workshops}, pp.\  224--236, 2018.

\bibitem[Feng et~al.(2023)Feng, Yang, Guo, and Li]{feng2023cvrecon}
Ziyue Feng, Leon Yang, Pengsheng Guo, and Bing Li.
\newblock Cvrecon: Rethinking 3d geometric feature learning for neural reconstruction.
\newblock \emph{arXiv preprint arXiv:2304.14633}, 2023.

\bibitem[Fu et~al.(2022)Fu, Zhang, Chen, Lu, Zhu, Zhou, Geiger, and Liao]{fu2022panoptic}
Xiao Fu, Shangzhan Zhang, Tianrun Chen, Yichong Lu, Lanyun Zhu, Xiaowei Zhou, Andreas Geiger, and Yiyi Liao.
\newblock Panoptic nerf: 3d-to-2d label transfer for panoptic urban scene segmentation.
\newblock In \emph{2022 International Conference on 3D Vision (3DV)}, pp.\  1--11. IEEE, 2022.

\bibitem[Furukawa \& Ponce(2009)Furukawa and Ponce]{furukawa2009accurate}
Yasutaka Furukawa and Jean Ponce.
\newblock Accurate, dense, and robust multiview stereopsis.
\newblock \emph{IEEE transactions on pattern analysis and machine intelligence}, 32\penalty0 (8):\penalty0 1362--1376, 2009.

\bibitem[Furukawa et~al.(2015)Furukawa, Hern{\'a}ndez, et~al.]{furukawa2015multi}
Yasutaka Furukawa, Carlos Hern{\'a}ndez, et~al.
\newblock Multi-view stereo: A tutorial.
\newblock \emph{Foundations and Trends{\textregistered} in Computer Graphics and Vision}, 9\penalty0 (1-2):\penalty0 1--148, 2015.

\bibitem[Gao et~al.(2023)Gao, Su, Liang, Yue, Yang, and Fu]{gao2023mc}
Yu~Gao, Lutong Su, Hao Liang, Yufeng Yue, Yi~Yang, and Mengyin Fu.
\newblock Mc-nerf: Muti-camera neural radiance fields for muti-camera image acquisition systems.
\newblock \emph{arXiv preprint arXiv:2309.07846}, 2023.

\bibitem[Gu(2023)]{zipcode}
Chun Gu.
\newblock Zipnerf-pytorch, 2023.
\newblock URL \url{https://github.com/SuLvXiangXin/zipnerf-pytorch}.
\newblock Accessed: 2023-04-23.

\bibitem[Guizilini et~al.(2020)Guizilini, Ambrus, Pillai, Raventos, and Gaidon]{guizilini20203d}
Vitor Guizilini, Rares Ambrus, Sudeep Pillai, Allan Raventos, and Adrien Gaidon.
\newblock 3d packing for self-supervised monocular depth estimation.
\newblock In \emph{Proceedings of the IEEE/CVF conference on computer vision and pattern recognition}, pp.\  2485--2494, 2020.

\bibitem[Jeong et~al.(2021)Jeong, Ahn, Choy, Anandkumar, Cho, and Park]{jeong2021self}
Yoonwoo Jeong, Seokjun Ahn, Christopher Choy, Anima Anandkumar, Minsu Cho, and Jaesik Park.
\newblock Self-calibrating neural radiance fields.
\newblock In \emph{Proceedings of the IEEE/CVF International Conference on Computer Vision}, pp.\  5846--5854, 2021.

\bibitem[Li et~al.(2023)Li, Wang, Yang, Liu, Qiu, and Wang]{li2023nerf}
Peihao Li, Shaohui Wang, Chen Yang, Bingbing Liu, Weichao Qiu, and Haoqian Wang.
\newblock Nerf-ms: Neural radiance fields with multi-sequence.
\newblock In \emph{Proceedings of the IEEE/CVF International Conference on Computer Vision}, pp.\  18591--18600, 2023.

\bibitem[Lin et~al.(2021)Lin, Ma, Torralba, and Lucey]{lin2021barf}
Chen-Hsuan Lin, Wei-Chiu Ma, Antonio Torralba, and Simon Lucey.
\newblock Barf: Bundle-adjusting neural radiance fields.
\newblock In \emph{Proceedings of the IEEE/CVF International Conference on Computer Vision}, pp.\  5741--5751, 2021.

\bibitem[Liu et~al.(2023)Liu, Kumar, Gu, Timofte, and Van~Gool]{liu2022va}
Ce~Liu, Suryansh Kumar, Shuhang Gu, Radu Timofte, and Luc Van~Gool.
\newblock Va-depthnet: A variational approach to single image depth prediction.
\newblock In \emph{The Eleventh International Conference on Learning Representations}, 2023.

\bibitem[Long et~al.(2020)Long, Liu, Theobalt, and Wang]{long2020occlusion}
Xiaoxiao Long, Lingjie Liu, Christian Theobalt, and Wenping Wang.
\newblock Occlusion-aware depth estimation with adaptive normal constraints.
\newblock In \emph{Computer Vision--ECCV 2020: 16th European Conference, Glasgow, UK, August 23--28, 2020, Proceedings, Part IX 16}, pp.\  640--657. Springer, 2020.

\bibitem[Long et~al.(2021)Long, Liu, Li, Theobalt, and Wang]{long2021multi}
Xiaoxiao Long, Lingjie Liu, Wei Li, Christian Theobalt, and Wenping Wang.
\newblock Multi-view depth estimation using epipolar spatio-temporal networks.
\newblock In \emph{Proceedings of the IEEE/CVF Conference on Computer Vision and Pattern Recognition}, pp.\  8258--8267, 2021.

\bibitem[Ma et~al.(2022)Ma, Teed, and Deng]{ma2022multiview}
Zeyu Ma, Zachary Teed, and Jia Deng.
\newblock Multiview stereo with cascaded epipolar raft.
\newblock In \emph{European Conference on Computer Vision}, pp.\  734--750. Springer, 2022.

\bibitem[Martin-Brualla et~al.(2021)Martin-Brualla, Radwan, Sajjadi, Barron, Dosovitskiy, and Duckworth]{martin2021nerf}
Ricardo Martin-Brualla, Noha Radwan, Mehdi~SM Sajjadi, Jonathan~T Barron, Alexey Dosovitskiy, and Daniel Duckworth.
\newblock Nerf in the wild: Neural radiance fields for unconstrained photo collections.
\newblock In \emph{Proceedings of the IEEE/CVF Conference on Computer Vision and Pattern Recognition}, pp.\  7210--7219, 2021.

\bibitem[Mei et~al.(2022)Mei, Zhu, Yan, Yan, Qiao, Chen, and Kretzschmar]{mei2022waymo}
Jieru Mei, Alex~Zihao Zhu, Xinchen Yan, Hang Yan, Siyuan Qiao, Liang-Chieh Chen, and Henrik Kretzschmar.
\newblock Waymo open dataset: Panoramic video panoptic segmentation.
\newblock In \emph{European Conference on Computer Vision}, pp.\  53--72. Springer, 2022.

\bibitem[Mildenhall et~al.(2021)Mildenhall, Srinivasan, Tancik, Barron, Ramamoorthi, and Ng]{mildenhall2021nerf}
Ben Mildenhall, Pratul~P Srinivasan, Matthew Tancik, Jonathan~T Barron, Ravi Ramamoorthi, and Ren Ng.
\newblock Nerf: Representing scenes as neural radiance fields for view synthesis.
\newblock \emph{Communications of the ACM}, 65\penalty0 (1):\penalty0 99--106, 2021.

\bibitem[M{\"u}ller et~al.(2022)M{\"u}ller, Evans, Schied, and Keller]{muller2022instant}
Thomas M{\"u}ller, Alex Evans, Christoph Schied, and Alexander Keller.
\newblock Instant neural graphics primitives with a multiresolution hash encoding.
\newblock \emph{ACM Transactions on Graphics (ToG)}, 41\penalty0 (4):\penalty0 1--15, 2022.

\bibitem[Pang et~al.(2023)Pang, Li, Tokmakov, Chen, Zagoruyko, and Wang]{pang2023standing}
Ziqi Pang, Jie Li, Pavel Tokmakov, Dian Chen, Sergey Zagoruyko, and Yu-Xiong Wang.
\newblock Standing between past and future: Spatio-temporal modeling for multi-camera 3d multi-object tracking.
\newblock In \emph{Proceedings of the IEEE/CVF Conference on Computer Vision and Pattern Recognition}, pp.\  17928--17938, 2023.

\bibitem[Rematas et~al.(2022)Rematas, Liu, Srinivasan, Barron, Tagliasacchi, Funkhouser, and Ferrari]{rematas2022urban}
Konstantinos Rematas, Andrew Liu, Pratul~P Srinivasan, Jonathan~T Barron, Andrea Tagliasacchi, Thomas Funkhouser, and Vittorio Ferrari.
\newblock Urban radiance fields.
\newblock In \emph{Proceedings of the IEEE/CVF Conference on Computer Vision and Pattern Recognition}, pp.\  12932--12942, 2022.

\bibitem[Sch{\"o}nberger et~al.(2016)Sch{\"o}nberger, Zheng, Frahm, and Pollefeys]{schonberger2016pixelwise}
Johannes~L Sch{\"o}nberger, Enliang Zheng, Jan-Michael Frahm, and Marc Pollefeys.
\newblock Pixelwise view selection for unstructured multi-view stereo.
\newblock In \emph{Computer Vision--ECCV 2016: 14th European Conference, Amsterdam, The Netherlands, October 11-14, 2016, Proceedings, Part III 14}, pp.\  501--518. Springer, 2016.

\bibitem[Shi et~al.(2022)Shi, Rong, Ni, Chen, and Zhang]{shi2022garf}
Yue Shi, Dingyi Rong, Bingbing Ni, Chang Chen, and Wenjun Zhang.
\newblock Garf: Geometry-aware generalized neural radiance field.
\newblock \emph{arXiv preprint arXiv:2212.02280}, 2022.

\bibitem[Sun et~al.(2020)Sun, Kretzschmar, Dotiwalla, Chouard, Patnaik, Tsui, Guo, Zhou, Chai, Caine, et~al.]{sun2020scalability}
Pei Sun, Henrik Kretzschmar, Xerxes Dotiwalla, Aurelien Chouard, Vijaysai Patnaik, Paul Tsui, James Guo, Yin Zhou, Yuning Chai, Benjamin Caine, et~al.
\newblock Scalability in perception for autonomous driving: Waymo open dataset.
\newblock In \emph{Proceedings of the IEEE/CVF conference on computer vision and pattern recognition}, pp.\  2446--2454, 2020.

\bibitem[Tancik et~al.(2022)Tancik, Casser, Yan, Pradhan, Mildenhall, Srinivasan, Barron, and Kretzschmar]{tancik2022block}
Matthew Tancik, Vincent Casser, Xinchen Yan, Sabeek Pradhan, Ben Mildenhall, Pratul~P Srinivasan, Jonathan~T Barron, and Henrik Kretzschmar.
\newblock Block-nerf: Scalable large scene neural view synthesis.
\newblock In \emph{Proceedings of the IEEE/CVF Conference on Computer Vision and Pattern Recognition}, pp.\  8248--8258, 2022.

\bibitem[Tosi et~al.(2023)Tosi, Tonioni, De~Gregorio, and Poggi]{tosi2023nerf}
Fabio Tosi, Alessio Tonioni, Daniele De~Gregorio, and Matteo Poggi.
\newblock Nerf-supervised deep stereo.
\newblock In \emph{Proceedings of the IEEE/CVF Conference on Computer Vision and Pattern Recognition}, pp.\  855--866, 2023.

\bibitem[Truong et~al.(2023)Truong, Rakotosaona, Manhardt, and Tombari]{truong2023sparf}
Prune Truong, Marie-Julie Rakotosaona, Fabian Manhardt, and Federico Tombari.
\newblock Sparf: Neural radiance fields from sparse and noisy poses.
\newblock In \emph{Proceedings of the IEEE/CVF Conference on Computer Vision and Pattern Recognition}, pp.\  4190--4200, 2023.

\bibitem[Turki et~al.(2022)Turki, Ramanan, and Satyanarayanan]{turki2022mega}
Haithem Turki, Deva Ramanan, and Mahadev Satyanarayanan.
\newblock Mega-nerf: Scalable construction of large-scale nerfs for virtual fly-throughs.
\newblock In \emph{Proceedings of the IEEE/CVF Conference on Computer Vision and Pattern Recognition}, pp.\  12922--12931, 2022.

\bibitem[Turki et~al.(2023)Turki, Zhang, Ferroni, and Ramanan]{turki2023suds}
Haithem Turki, Jason~Y Zhang, Francesco Ferroni, and Deva Ramanan.
\newblock Suds: Scalable urban dynamic scenes.
\newblock In \emph{Proceedings of the IEEE/CVF Conference on Computer Vision and Pattern Recognition}, pp.\  12375--12385, 2023.

\bibitem[Vakalopoulou et~al.(2018)Vakalopoulou, Chassagnon, Bus, Marini, Zacharaki, Revel, and Paragios]{vakalopoulou2018atlasnet}
Maria Vakalopoulou, Guillaume Chassagnon, Norbert Bus, Rafael Marini, Evangelia~I Zacharaki, M-P Revel, and Nikos Paragios.
\newblock Atlasnet: Multi-atlas non-linear deep networks for medical image segmentation.
\newblock In \emph{Medical Image Computing and Computer Assisted Intervention--MICCAI 2018: 21st International Conference, Granada, Spain, September 16-20, 2018, Proceedings, Part IV 11}, pp.\  658--666. Springer, 2018.

\bibitem[Wang et~al.(2023{\natexlab{a}})Wang, Liu, Chen, Liu, Liu, Komura, Theobalt, and Wang]{wang2023f2}
Peng Wang, Yuan Liu, Zhaoxi Chen, Lingjie Liu, Ziwei Liu, Taku Komura, Christian Theobalt, and Wenping Wang.
\newblock F2-nerf: Fast neural radiance field training with free camera trajectories.
\newblock In \emph{Proceedings of the IEEE/CVF Conference on Computer Vision and Pattern Recognition}, pp.\  4150--4159, 2023{\natexlab{a}}.

\bibitem[Wang et~al.(2023{\natexlab{b}})Wang, Shen, Gao, Huang, Munkberg, Hasselgren, Gojcic, Chen, and Fidler]{wang2023neural}
Zian Wang, Tianchang Shen, Jun Gao, Shengyu Huang, Jacob Munkberg, Jon Hasselgren, Zan Gojcic, Wenzheng Chen, and Sanja Fidler.
\newblock Neural fields meet explicit geometric representations for inverse rendering of urban scenes.
\newblock In \emph{Proceedings of the IEEE/CVF Conference on Computer Vision and Pattern Recognition}, pp.\  8370--8380, 2023{\natexlab{b}}.

\bibitem[Wang et~al.(2021)Wang, Wu, Xie, Chen, and Prisacariu]{wang2021nerf}
Zirui Wang, Shangzhe Wu, Weidi Xie, Min Chen, and Victor~Adrian Prisacariu.
\newblock Nerf--: Neural radiance fields without known camera parameters.
\newblock \emph{arXiv preprint arXiv:2102.07064}, 2021.

\bibitem[Xia et~al.(2022)Xia, Tang, Timofte, and Van~Gool]{xia2022sinerf}
Yitong Xia, Hao Tang, Radu Timofte, and Luc Van~Gool.
\newblock Sinerf: Sinusoidal neural radiance fields for joint pose estimation and scene reconstruction.
\newblock \emph{arXiv preprint arXiv:2210.04553}, 2022.

\bibitem[Xie et~al.(2023)Xie, Zhang, Li, Zhang, and Zhang]{xie2022s}
Ziyang Xie, Junge Zhang, Wenye Li, Feihu Zhang, and Li~Zhang.
\newblock S-nerf: Neural radiance fields for street views.
\newblock In \emph{The Eleventh International Conference on Learning Representations}, 2023.

\bibitem[Yang et~al.(2023)Yang, Chen, Wang, Manivasagam, Ma, Yang, and Urtasun]{yang2023unisim}
Ze~Yang, Yun Chen, Jingkang Wang, Sivabalan Manivasagam, Wei-Chiu Ma, Anqi~Joyce Yang, and Raquel Urtasun.
\newblock Unisim: A neural closed-loop sensor simulator.
\newblock In \emph{Proceedings of the IEEE/CVF Conference on Computer Vision and Pattern Recognition}, pp.\  1389--1399, 2023.

\bibitem[Yao et~al.(2018)Yao, Luo, Li, Fang, and Quan]{yao2018mvsnet}
Yao Yao, Zixin Luo, Shiwei Li, Tian Fang, and Long Quan.
\newblock Mvsnet: Depth inference for unstructured multi-view stereo.
\newblock In \emph{Proceedings of the European conference on computer vision (ECCV)}, pp.\  767--783, 2018.

\bibitem[Yin et~al.(2022)Yin, Liu, Shen, Hengel, and Sun]{yin2022devil}
Wei Yin, Yifan Liu, Chunhua Shen, Anton van~den Hengel, and Baichuan Sun.
\newblock The devil is in the labels: Semantic segmentation from sentences.
\newblock \emph{arXiv preprint arXiv:2202.02002}, 2022.

\bibitem[Zhang et~al.(2020)Zhang, Riegler, Snavely, and Koltun]{zhang2020nerf++}
Kai Zhang, Gernot Riegler, Noah Snavely, and Vladlen Koltun.
\newblock Nerf++: Analyzing and improving neural radiance fields.
\newblock \emph{arXiv preprint arXiv:2010.07492}, 2020.

\bibitem[Zhang et~al.(2018)Zhang, Isola, Efros, Shechtman, and Wang]{zhang2018unreasonable}
Richard Zhang, Phillip Isola, Alexei~A Efros, Eli Shechtman, and Oliver Wang.
\newblock The unreasonable effectiveness of deep features as a perceptual metric.
\newblock In \emph{Proceedings of the IEEE conference on computer vision and pattern recognition}, pp.\  586--595, 2018.

\bibitem[Zhang et~al.(2023)Zhang, Kundu, Funkhouser, Guibas, Su, and Genova]{zhang2023nerflets}
Xiaoshuai Zhang, Abhijit Kundu, Thomas Funkhouser, Leonidas Guibas, Hao Su, and Kyle Genova.
\newblock Nerflets: Local radiance fields for efficient structure-aware 3d scene representation from 2d supervision.
\newblock In \emph{Proceedings of the IEEE/CVF Conference on Computer Vision and Pattern Recognition}, pp.\  8274--8284, 2023.

\end{thebibliography}
\bibliographystyle{conference}

\appendix
\section{Appendix}
\subsection{Implementation Details}
We train our UC-NeRF for 40k iterations using Adam optimizer with a batch size of $32384$. The learning rate is logarithmically reduced from $0.008$ to $0.001$, with a warm-up phase consisting of $5000$ iterations. The training takes about $3$ hours for a scene with about $300$ images on two V100 GPUs.

% \paragraph{Sky Modeling} 
% Following Urban NeRF, we use a coordinate-based neural
% network to provide directional

\paragraph{Layer-based Color Correction}
% latent code dimension; MLP layers; L1 normalization; sky and foreground
In UC-NeRF, we model a scene as the foreground and sky. The foreground is represented by Zip-NeRF while the sky is modeled by the vanilla NeRF (~\cite{mildenhall2021nerf}). The weight of sky loss is set to $2 \times 10^{-3}$.
The dimension of sky latent code and foreground latent code is set to 4. For the MLP that decodes the latent code, we use three layers with 256 hidden units. The weight of transformation regularization is set to $2 \times 10^{-3}$.

\begin{figure*}
\centering    
\includegraphics[width=\linewidth]{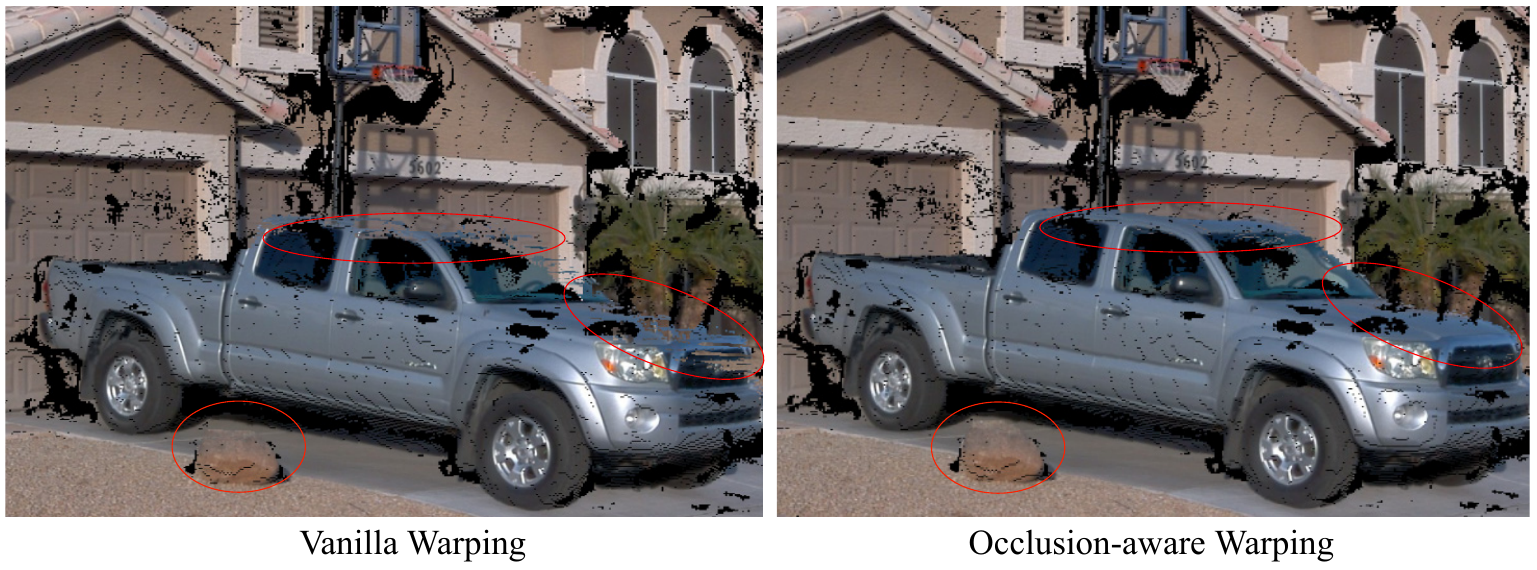}
\caption{Occlusion problem in warping.} 
\label{fig: occlusion}
\end{figure*}

\paragraph{Virtual Warping}
%occlusion; filter threshold; ray batch percentage for rendering; randomly generate;
For virtual warping,
% each virtual pose is generated by adding the perturbation with a rotation between $0-20$ degrees and a translation between $0-1$ m for a known pose.
% The directions of rotation and translation are randomly selected from the positive and negative directions of the three coordinate axes in the camera coordinate system. 
we randomly sample $9$ virtual poses for each existing pose.
The occlusion needs to be considered in the case of multiple pixels of the known view warping to the same pixel in the virtual view,  which is shown in the red boxes of Fig.~\ref{fig: occlusion}. We resolve this conflict by taking the warped pixel with the smallest depth value. 
For generating the geometric consistency mask, we set $6$ target views to check the depth consistency. Only pixels with depth absolute relative error within the range of $0.01$ for at least $4$ neighboring views are retained. 
For each training batch, we sample the rays from the real and virtual images at a ratio of $4:1$ respectively. 

\paragraph{Spatiotemporally Constrained Pose Refinement}
% correspondence by Superpoint; threshold for correspondence filter; 
We use the reprojection error to optimize the camera poses. To calculate the reprojection error, the feature points need to be extracted from images. We use Superpoint (~\cite{detone2018superpoint}) to detect and describe the keypoints. The keypoints are matched by mutual nearest neighbors and the confidence higher than $0.95$ is preserved. For each view, we match it with the subsequent ten frames captured by the same camera and the subsequent twenty frames from different cameras.
Image pairs with more than $30$ matching points are retained.

% \paragraph{Selected Scene Indexes}

\subsection{Special case for spatiotemporal constraint}
In Sec.~\ref{sec:pr}, we model the pose of $kth$ camera at timestamp $i$ as $\mathbf{T}^{i}_{k} = \mathbf{T}^{i} \Delta \mathbf{T}_{k}$. Here pose refers to the transformation from the camera coordinate to the world coordinate. $\mathbf{T}^{i}$ refers to the car's ego pose at time $i$. $\Delta \mathbf{T}_{k}$ refers to the transformation from the $kth$ camera's coordinate to the ego coordinate, which is temporally consistent. During bundle adjustment, $\mathbf{T}^{i}_{k}$, and $\mathbf{T}^{j}_ {l}$ are optimized as Eq.~\ref{eq:ba} in Sec.~\ref{sec:pr}, where $(\mathbf{T}^{j}_{l})^{-1}\mathbf{T}^{i}_{k}$ are expressed as:

\begin{equation}
\begin{aligned}
(\mathbf{T}^{j}_{l})^{-1} \mathbf{T}^{i}_{ k} & =\left(\mathbf{T}^j \Delta \mathbf{T}_l\right)^{-1} \mathbf{T}^i \Delta \mathbf{T}_k \\
& =\Delta \mathbf{T}_l^{-1} (\mathbf{T}^j)^{-1} \mathbf{T}^i \Delta \mathbf{T}_k
\end{aligned}.
\end{equation}

Expressing $(\mathbf{T}^{j})^{-1} \mathbf{T}^{i}=\left[\begin{array}{cc}
\mathbf{R}^{i,j} & \mathbf{t}^{i,j} \\
\mathbf{0} & \mathbf{1}
\end{array}\right]$, $\Delta \mathbf{T}_{l} = \left[\begin{array}{cc}
\Delta \mathbf{R}_{l} & \Delta \mathbf{t}_{l} \\
\mathbf{0} & \mathbf{1}
\end{array}\right]$, and $\Delta \mathbf{T}_{k} = \left[\begin{array}{cc}
\Delta \mathbf{R}_{k} & \Delta \mathbf{t}_{k} \\
\mathbf{0} & \mathbf{1}
\end{array}\right]$, then

\begin{equation}
\begin{aligned}
\label{eq:relativetrans}
(\mathbf{T}^{j}_{l})^{-1} \mathbf{T}^{i}_{k} & =\left[\begin{array}{cc}
\Delta \mathbf{R}_l^{\top} & -\Delta \mathbf{R}_l^{\top} \Delta \mathbf{t}_l \\
\mathbf{0} & \mathbf{1}
\end{array}\right]\left[\begin{array}{cc}
\mathbf{R}^{i,j} & \mathbf{t}^{i,j} \\
\mathbf{0} & \mathbf{1}
\end{array}\right]\left[\begin{array}{cc}
\Delta \mathbf{R}_k & \Delta \mathbf{t}_k \\
\mathbf{0} & \mathbf{1}
\end{array}\right] \\
& = \left[\begin{array}{cc}
\Delta \mathbf{R}_l^{\top} \mathbf{R}^{i,j} \Delta \mathbf{R}_k & \Delta \mathbf{R}_l^{\top} \mathbf{R}^{i,j} \Delta \mathbf{t}_k+\Delta \mathbf{R}_l^{\top} \mathbf{t}^{i,j}-\Delta \mathbf{R}^{\top}_{l}\ \Delta \mathbf{t}_l \\
\mathbf{0} & \mathbf{1}
\end{array}\right] .
\end{aligned}
\end{equation}

% \left[\begin{array}{cc}
% \Delta \mathbf{R}_k^{\top} \mathbf{R} \Delta \mathbf{R}_k & \Delta \mathbf{R}_k^{\top} \mathbf{R} \Delta \mathbf{t}_k+\Delta \mathbf{R}_k^{\top} \mathbf{t}-\Delta \mathbf{R}_k^{\top} \Delta \mathbf{t}_k \\
% \mathbf{0} & \mathbf{1}
% \end{array}\right]
% \end{aligned}.
% \end{equation}

When the vehicle is moving straight without any rotation, $\mathbf{R}^{i,j}$ equals to the identity matrix. Thus, the Eq.~\ref{eq:relativetrans} is simplified as:
\begin{equation}
(\mathbf{T}^{j}_{l})^{-1} \mathbf{T}^{i}_{k}=\left[\begin{array}{cc}
\Delta \mathbf{R}_l^{\top} \Delta \mathbf{R}_k & \Delta \mathbf{R}_l^{\top} \Delta \mathbf{t}_k+\Delta \mathbf{R}_l^{\top} \mathbf{t}^{i,j}-\Delta \mathbf{R}^{\top}_{l}\ \Delta \mathbf{t}_l \\
\mathbf{0} & \mathbf{1}
\end{array}\right].
\label{eq:simp1}
\end{equation}

If the image correspondences are not established across cameras, $i.e.$, $l=k$, then Eq.~\ref{eq:simp1} can further simplified as:

\begin{equation}
\label{method:simpletrans}
(\mathbf{T}^{j}_{l})^{-1} \mathbf{T}^{i}_{k}=\left[\begin{array}{cc}
\mathbf{I} & \Delta \mathbf{R}^{\top}_{k} \mathbf{t}^{i,j} \\
\mathbf{0} & \mathbf{1}
\end{array}\right],
\end{equation}

which suggests that the relative transformation between any two neighboring poses of the same camera $k$ remains unaffected by $\Delta \mathbf{t}_{k}$, thus resulting in a lack of constraint on the camera's translation $\Delta \mathbf{t}_{k}$ during the optimization process. This implies that the image correspondences across both cameras and timestamps ensure a robust constraint on inter-camera transformation.

% Utilizing the same derivation method, $\mathbf{T}_{j, l}^{-1} \mathbf{T}_{i, k}$ can be expressed as:

% \begin{equation}
% \mathbf{T}_{j, l}^{-1} \mathbf{T}_{i, k}=\left[\begin{array}{cc}
% \Delta \mathbf{R}_l^{\top} \mathbf{R} \Delta \mathbf{R}_k & \Delta \mathbf{R}_l^{\top} \mathbf{R} \Delta \mathbf{t}_k+\Delta \mathbf{R}_l^{\top} \mathbf{t}-\Delta \mathbf{R}^{\top}_{l}\ \Delta \mathbf{t}_l \\
% \mathbf{0} & \mathbf{1}
% \end{array}\right] .
% \end{equation}

% Even in the case of non-rotational motion where $\mathbf{R} = \mathbf{I}$, 

% \begin{equation}
% \mathbf{T}_{j, l}^{-1} \mathbf{T}_{i, k}=\left[\begin{array}{cc}
% \Delta \mathbf{R}_l^{\top} \Delta \mathbf{R}_k & \Delta \mathbf{R}_l^{\top} \Delta \mathbf{t}_k+\Delta \mathbf{R}_l^{\top} \mathbf{t}-\Delta \mathbf{R}^{\top}_{l}\ \Delta \mathbf{t}_l \\
% \mathbf{0} & \mathbf{1}
% \end{array}\right].
% \end{equation}

\subsection{Experiments}

\subsubsection{Application: Synthesized Views for Monocular Depth Estimation}

\begin{table}
\centering
\footnotesize
\caption{The accuracy of depth estimation using VA-DepthNet before and after adding our rendered novel views (VA-DepthNet$^*$) for training.}
\label{tab:depth}
\begin{tabular}{@{}l|ccc@{}}
\toprule
Method & Abs Rel $\downarrow$ & RMSE $\downarrow$ & $\delta1$ $\uparrow$ \\ \midrule
VA-DepthNet  & 0.078 & 2.82 & 93.7 \\
VA-DepthNet$^*$  & $\mathbf{0.076}$ & $\mathbf{2.64}$ &  $\mathbf{94.2}$ \\ \bottomrule
\end{tabular}
\end{table}

\begin{figure*}
\centering    
\includegraphics[width=\linewidth]{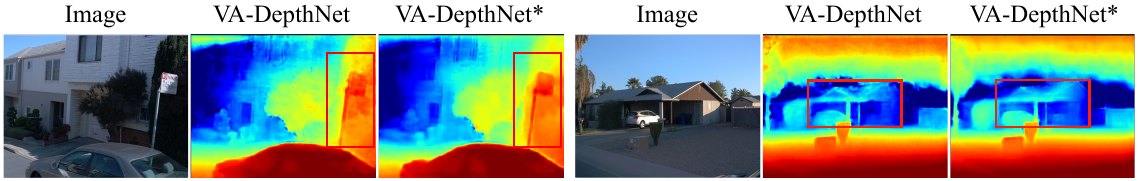}
\caption{Compared to training VA-DepthNet (\cite{liu2022va}) on the original data, augmenting it with our rendered novel views (VA-DepthNet$^{*}$) leads to improved depth estimation. Depth of VA-DepthNet$^{*}$ in the red boxes exhibits sharper edges and smoother surfaces.} 
\label{fig: depthimprove}
\end{figure*}

With the obtained 3D NeRF, we can generate additional photo-realistic images from novel viewpoints.
The synthesized images can facilitate downstream perception tasks like monocular depth estimation. 
We first train VA-DepthNet, a state-of-the-art monocular depth estimation model (\cite{liu2022va}), on the original real images.
We then train the model by combining the original real images and the new synthesized images (VA-DepthNet*).
As Tab.~\ref{tab:depth} illustrates, the accuracy of the estimated depth is improved with such a data augmentation. Fig.~\ref{fig: depthimprove} also shows such an operation leads to sharper edges and more accurate predictions.

\subsubsection{More Ablation Study results}

\begin{figure*}[b!]
\centering    
\includegraphics[width=\linewidth]{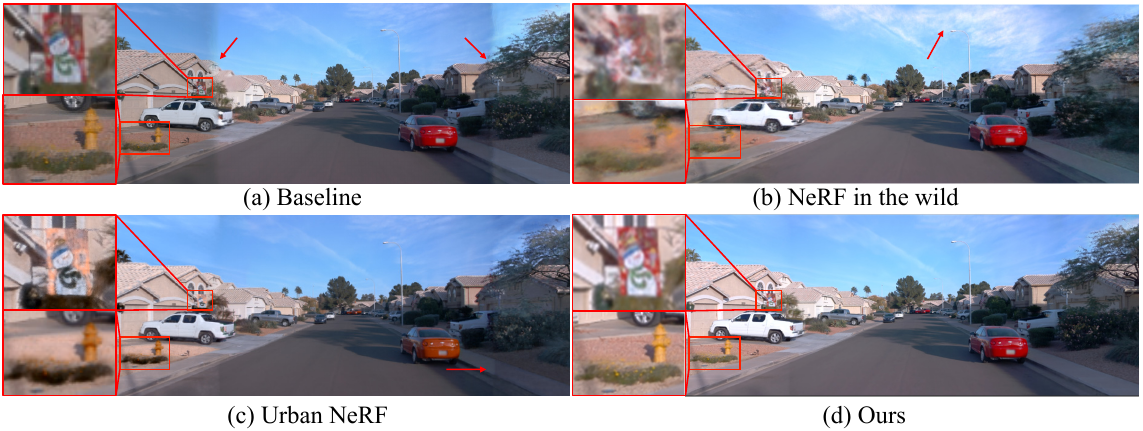}
\caption{Comparison of color correction strategies for rendering images with large field of view.} 
\label{fig: lcablation}
\end{figure*}

\paragraph{Color Correction Strategies.} 
\begin{table}[ht]
\centering
\footnotesize
\caption{Comparison of different strategies for color correction.}
\begin{tabular}{@{}l|lll@{}}
\toprule
Method & PSNR $\uparrow$ & SSIM $\uparrow$ & LPIPS $\downarrow$ \\ \midrule
NeRF in the wild (~\cite{martin2021nerf}) & 25.59 & 0.839 & 0.389  \\
% \mk{NeRF-MS} (~\cite{li2023nerf}) & 26.87 & 0.828 & 0.392 \\
Urban-NeRF (~\cite{rematas2022urban}) &  27.89 & 0.849 & 0.378 \\
Ours   & \textbf{28.15} & \textbf{0.851} & \textbf{0.374} \\ \bottomrule
\end{tabular}
\label{tab:lcablation}
\end{table} %

We compare our color correction with other strategies that also model the image-dependent appearance with latent codes, as done in NeRF in the wild (~\cite{martin2021nerf}) and Urban-NeRF (~\cite{rematas2022urban}). 
As shown in Fig.~\ref{fig: lcablation}, the baseline (a) exhibits noticeable color discontinuities in regions where camera views overlap (indicated by red arrows). Although NeRF in the wild (b) models image-dependent appearance through latent codes, the absence of constraints on latent codes results in the disentanglement of attributes not related to the cause of color inconsistency. As a result, the panoramic rendering produced by it exhibits significant blurriness on both sides (emphasized by red boxes), along with additional texture artifacts in the sky (red arrows).
Urban-NeRF (c) also decodes the latent code into affine transformations to model image-dependent appearance. However, due to the absence of separate modeling for color transformations across different image regions, color discontinuities, as indicated by the red arrows, persist in the overlapping regions of the cameras. Additionally, the lack of constraints for inter-camera color correction results in misalignment within the same color space across different regions. When employing a single latent code for rendering the panorama image, the color within the central region adheres closely to reality. However, severe
color deviations occur in the peripheral area, such as the grass in the red box, which appears black instead of its natural color.
In contrast, our method addresses the color inconsistencies and ensures clear details with consistent colors. Tab.~\ref{tab:lcablation} also demonstrates that our approach achieves the best rendering results.

\paragraph{Pose Refinement Strategies}
\label{abexp:pr}

\begin{figure*}
\centering    
\includegraphics[width=\linewidth]{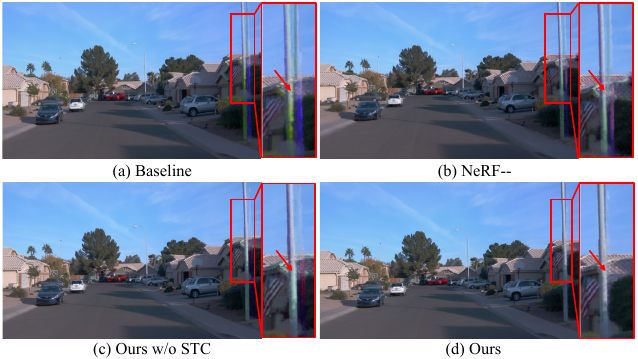}
\caption{Quantitative comparison of different pose optimization strategies. We effectively eliminate the ghosting of the pole. STC refers to the proposed spatiotemporal constraint.} 
\label{fig: poseablation}
\end{figure*}

S-NeRF (~\cite{xie2022s}) explores the performance of various NeRF algorithms that jointly optimize poses in urban scenes and found that NeRF$--$ (~\cite{wang2021nerf}) yields the best results. Thus, we compare our method with NeRF$--$ and validate the significance of the spatiotemporal constraint proposed in our paper.
As demonstrated in Tab.~\ref{tab: poseablation}, NeRF$--$ indeed enhances rendering results compared to the baseline which does not refine poses, and performance does not exhibit a significant change with the inclusion of the spatiotemporal constraint. In contrast, our spatiotemporally constrained pose refinement achieves a remarkable $238\%$ improvement compared to NeRF$--$. 
% Furthermore, it's crucial to note that the spatial constraint must be complemented by the spatial-temporal constraint; otherwise, performance experiences a noticeable decline due to the absence of constraints on inter-camera relative translations, as discussed in Sec.~\ref{sec:pr}. 
Fig.~\ref{fig: poseablation} clearly demonstrates how our method effectively resolves rendering artifacts.  
Due to color variation among different cameras, NeRF$--$ struggles to achieve precise pose optimization based on the photometric error. The ghosting of the pole (b) does not change significantly compared to the baseline (a). However, when poses are optimized by explicit pixel correspondences, the ghosting is noticeably reduced (c). Furthermore, with the addition of our spatiotemporal constraint, the artifact completely disappears (d).

\begin{table}
\centering
\footnotesize
\caption{Ablation study on different strategies for pose refinement. NeRF$--$ refine poses within the NeRF framework by photometric loss while we refine poses based on explicit pixel correspondences among images. STC refers to the proposed spatiotemporal constraint.}
\begin{tabular}{@{}l|lll@{}}
\toprule
Method & PSNR $\uparrow$ & SSIM $\uparrow$ & LPIPS $\downarrow$ \\ \midrule
Baseline & 28.09 & 0.851 & 0.374 \\
\midrule 
NeRF$--$ & 28.48 & 0.851 & 0.383 \\
NeRF$--$ w/ STC & 28.51 & 0.852 & 0.383 \\
\midrule
% Ours w/o STC & 26.94 & 0.865 & 0.398 \\
Ours w/o STC & 29.01 & 0.866 & 0.376 \\
% SC-NeRF &     &      &       \\
Ours & $\mathbf{29.14}$ & $\mathbf{0.867}$ & $\mathbf{0.355}$ \\ \bottomrule
\end{tabular}
\label{tab: poseablation}
\end{table}

\begin{table*}
  % \vspace{-0.5cm}
  \centering
  \footnotesize
  \caption{Comparison of diverse weather conditions. (Waymo Segment-100170, Waymo Segment-150908)}
    \newcolumntype{"}{@{\hskip\tabcolsep\vrule width 1.2pt\hskip\tabcolsep}}
  \begin{tabular}{@{}l|ccc"ccc@{}}
    \toprule
    & & Rainy  & & & Night & \\
    %\hline
    Method & PSNR $\uparrow$ & SSIM $\uparrow$ & LPIPS $\downarrow$ & PSNR $\uparrow$ & SSIM $\uparrow$ & LPIPS $\downarrow$ \\
  \hline
    Zip-NeRF[3] & 27.65 & 0.831 & 0.434 & 30.69 & 0.864 & 0.512 \\
    UC-NeRF (Ours) & \textbf{30.03} & \textbf{0.866} & \textbf{0.387} & \textbf{31.32} & \textbf{0.869} & \textbf{0.491} \\
    \bottomrule
  \end{tabular}
\label{tab:diverseweather}
\end{table*}

\begin{figure*}
\centering    
\includegraphics[width=\linewidth]{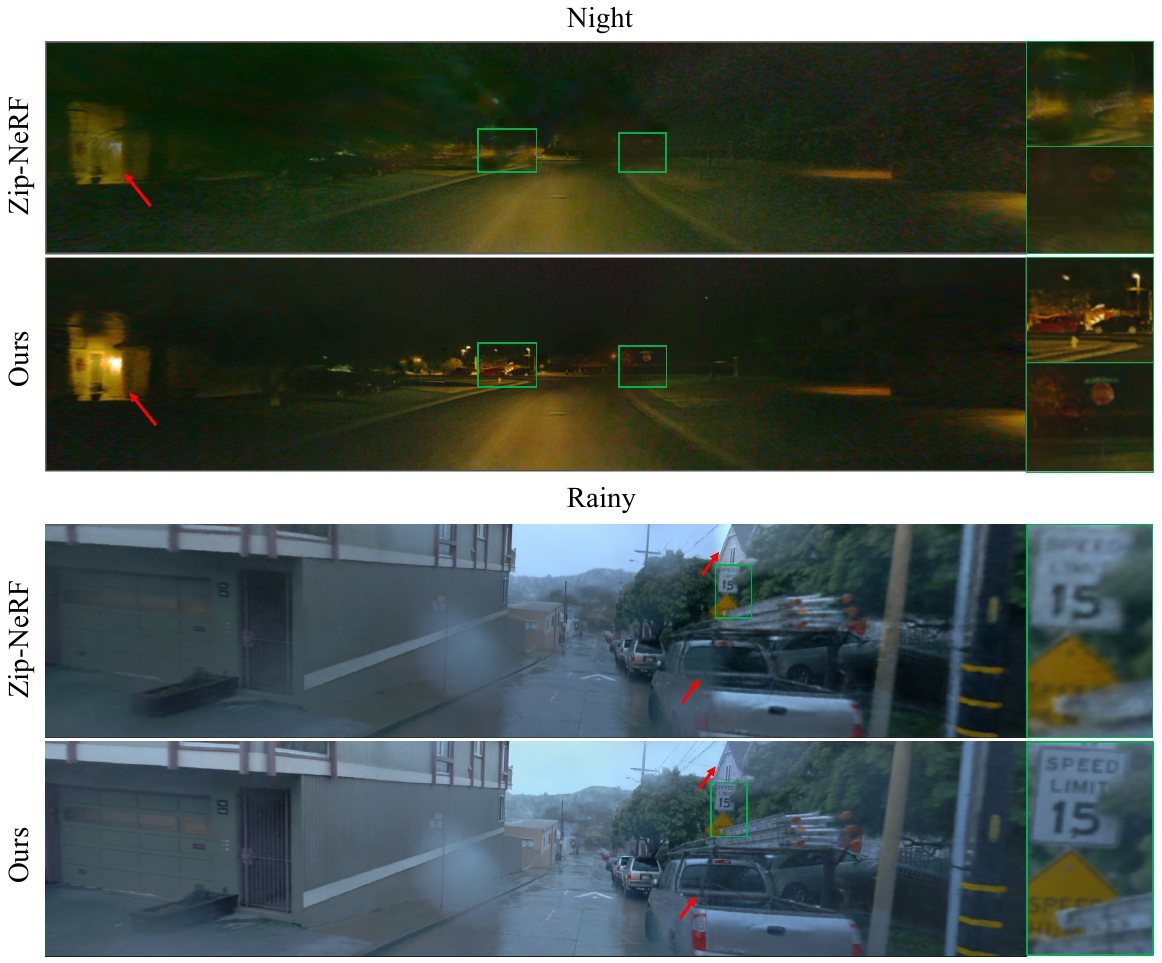}
\caption{$180^{\circ} $ panorama rendering results in night and rainy conditions.} 
\label{fig: diverseweather}
\vspace{-0.5cm}
\end{figure*}

\paragraph{Results on diverse weather conditions}
We further validate the robustness of our method under nighttime and rainy conditions. As shown in Tab.~\ref{tab:diverseweather}, our approach still significantly outperforms ZipNeRF in these scenarios. As illustrated in Fig.~\ref{fig: diverseweather}, we manage to eliminate the rendering artifacts caused by color inconsistency (indicated by red arrows) and achieve better rendering of details (highlighted by the green boxes).

\begin{table}
\centering
\footnotesize
\caption{Ablation study on the weight of sky loss.}
\begin{tabular}{@{}c|lll@{}}
\toprule
$w_{sky}$ &  PSNR $\uparrow$ & SSIM $\uparrow$ & LPIPS $\downarrow$ \\ \midrule
0 & 27.89 & 0.849 & 0.378 \\
0.001 & 28.05 & 0.851 & 0.379  \\
0.002 & \textbf{28.15} & \textbf{0.851}  &  \textbf{0.374}    \\
0.004 & 27.98 & 0.850 & 0.378 \\
%   & 27.92 & 0.849 & 0.378 \\
% 0 & 0.002
%   & 27.95 & 0.838 & 0.371 \\
%  0 & 0.004
%   & 27.81 & 0.845 & 0.381 \\ \midrule
%  0.002 & 0.002
%  &  \textbf{28.15} & \textbf{0.851} & \textbf{0.374} \\ 
 \bottomrule
\end{tabular}
\label{tab:skyloss}
\end{table}

\paragraph{Weight for Sky Loss}
As shown in Tab.~\ref{tab:skyloss}, we compare the performance of our UC-NeRF using different weights for sky loss.  Our UC-NeRF is not very sensitive to the changes in the loss weights. Using a large weight of sky loss might diminish the weight of the photometric loss, leading to a slight performance decline. $[0.001, 0.002]$ is the reasonable range for our loss weight.

\paragraph{More Results on Layer-based Color Correction}

\begin{figure*}[b!]
\centering    
\includegraphics[width=\linewidth]{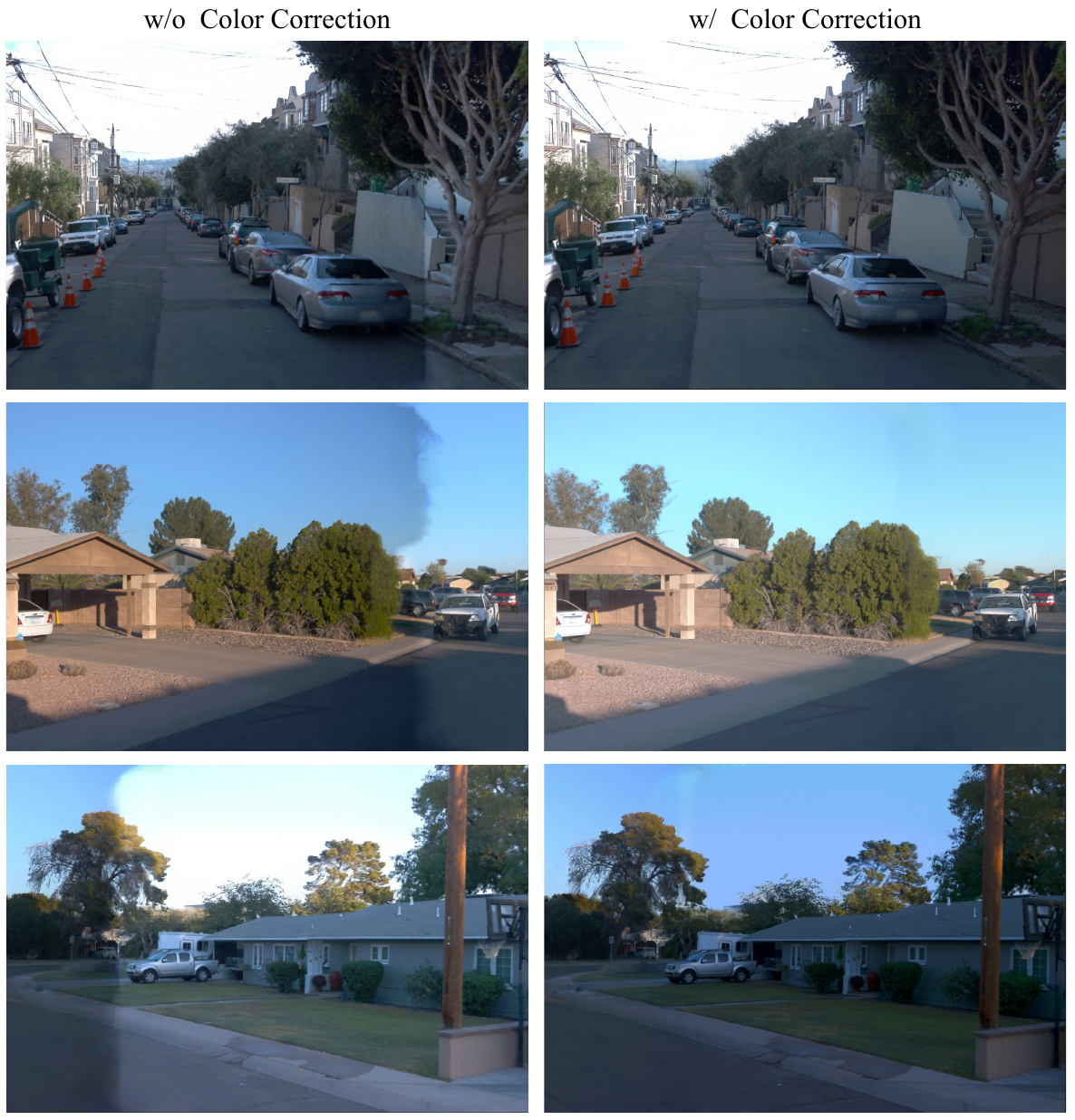}
\caption{More results on color correction.} 
\label{fig: lcexp}
\end{figure*}

As shown in Fig~\ref{fig: lcexp}, we present additional rendering results in overlap regions of different cameras. The original NeRF exhibits significant color inconsistencies in overlapping areas. Through color correction, we are able to render images that maintain global color consistency.

\paragraph{More Results on Spatiotemporally Constrained Pose Refinement}

\begin{figure*}
\centering    
\includegraphics[width=\linewidth]{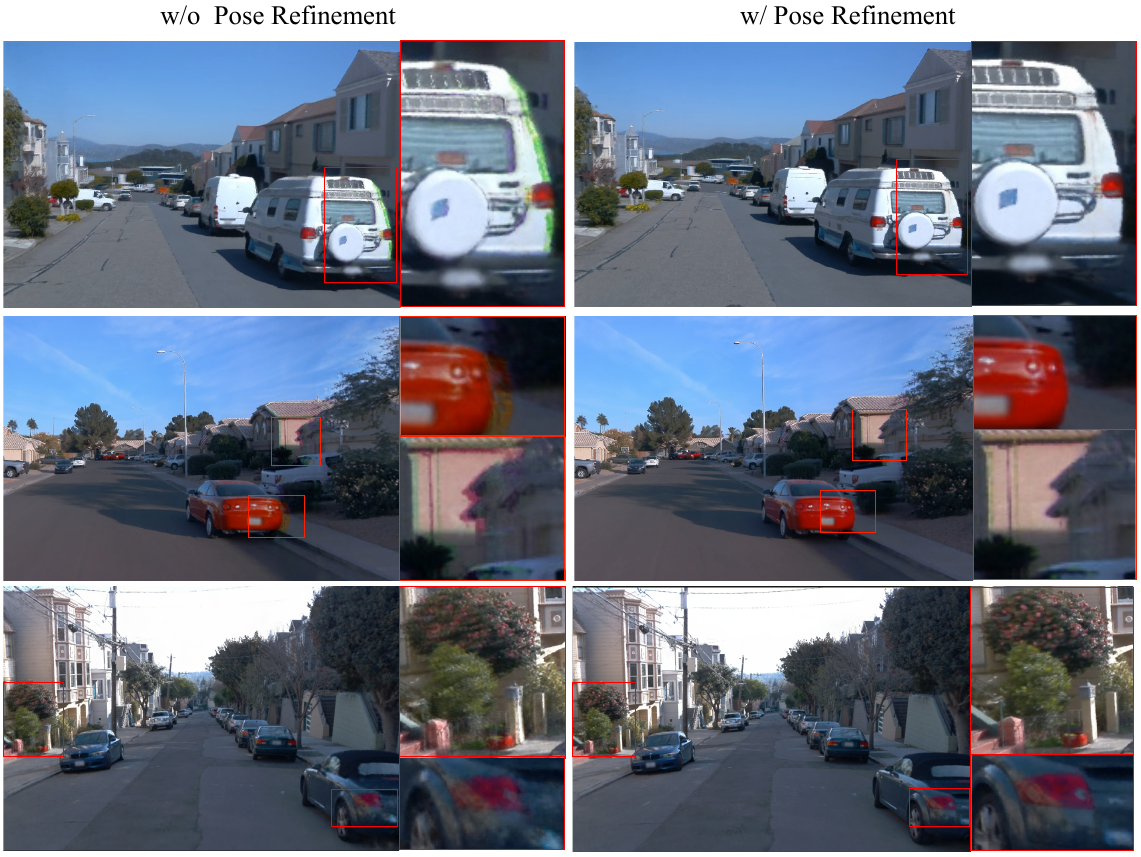}
\caption{More results on pose refinement.} 
\label{fig: posexp}
\end{figure*}

As shown in Fig~\ref{fig: posexp}, the issue of rendering artifacts and blurriness caused by pose errors is quite common in the multi-camera setup. Based on our observations, they typically occur in the overlapping regions captured by different cameras. This implies that these rendering artifacts result from errors in the relative transformations between the cameras. By explicitly modeling the relative transformations between cameras and ensuring spatiotemporal constraints during optimization, it is evident that these problems have been significantly addressed (shown in red boxes).

\subsubsection{More Results on Waymo and Nuscenes}

\begin{figure*}
\centering    
\vspace{-1.4cm}
\includegraphics[width=\linewidth]{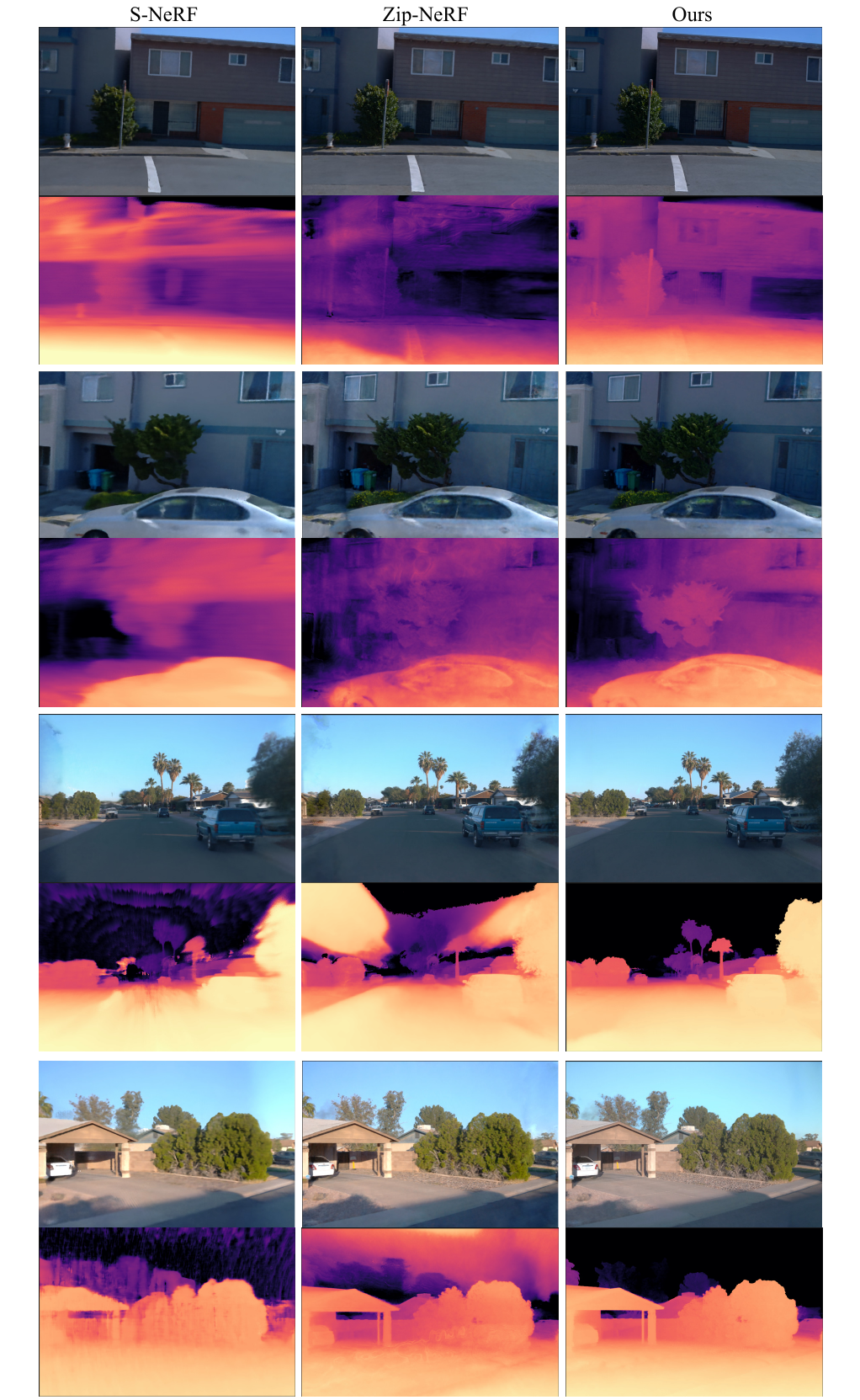}
\caption{Comparison of the rendering results with the state-of-the-art S-NeRF and Zip-NeRF in Waymo.} 
\label{fig: waymo}
\end{figure*}

\begin{figure*}
\centering    
\vspace{-1.4cm}
\includegraphics[width=\linewidth]{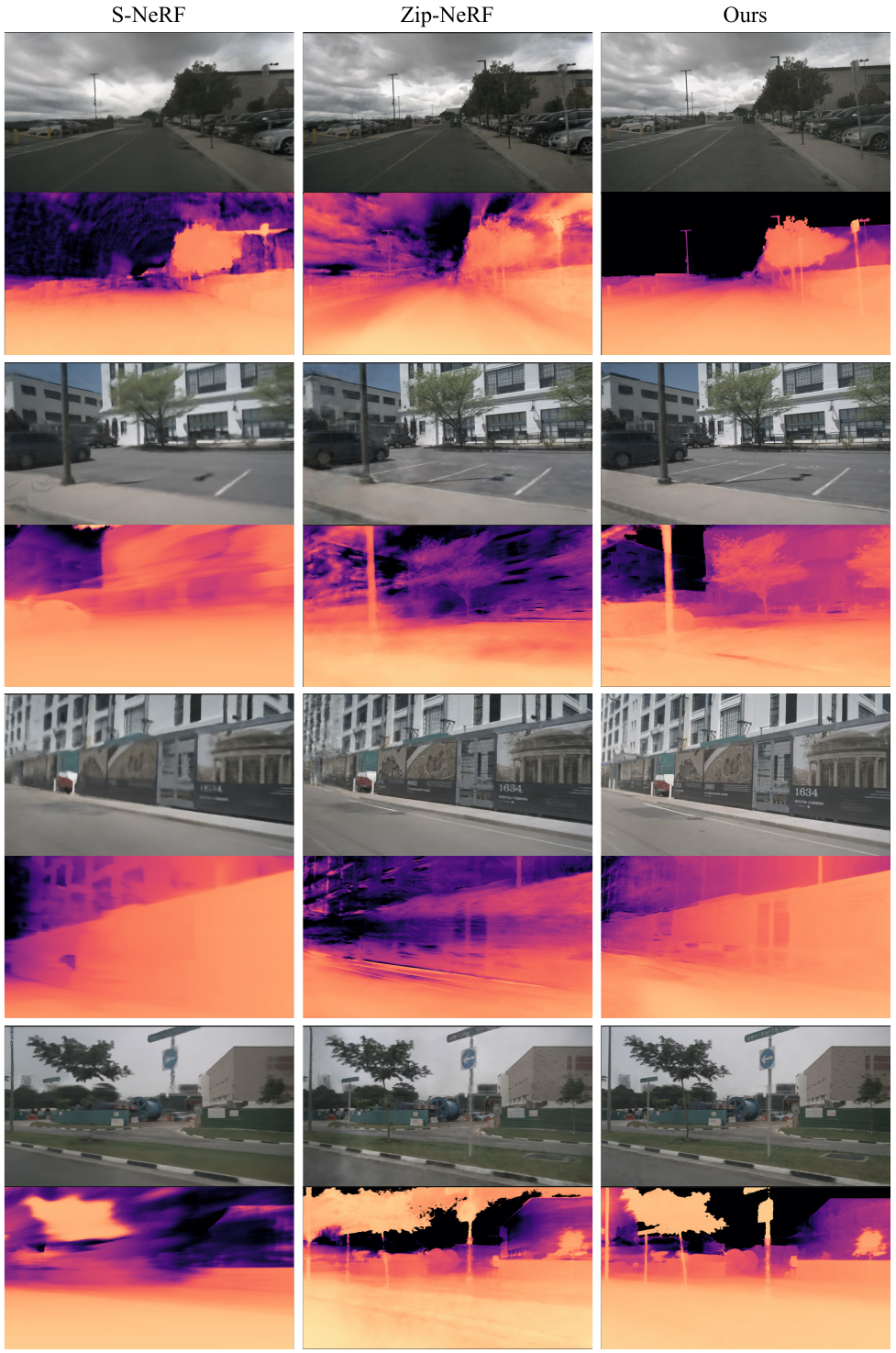}
\caption{Comparison of the rendering results and dept maps with S-NeRF~\cite{tosi2023nerf} (left) and Zip-NeRF~\cite{barron2023zip} (middle) in NuScenes.} 
\label{fig: nuscenes}
\end{figure*}

\begin{figure*}
\centering    
\includegraphics[width=\linewidth]{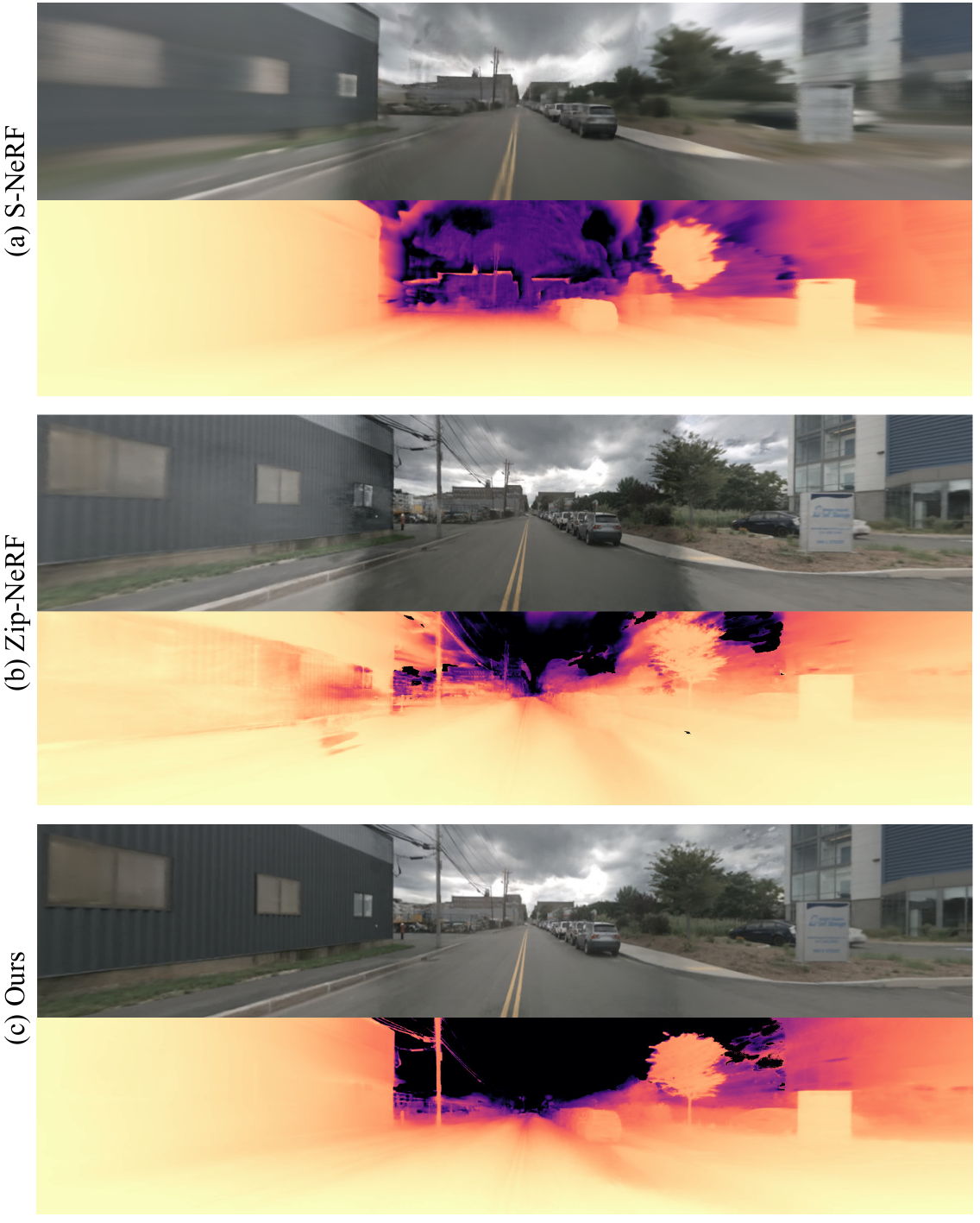}
\caption{$180^{\circ} $ panaroma rendering results in NuScenes.} 
\label{fig: nuscenesparanoma}
\end{figure*}

We further present rendering results on the Waymo and NuScenes datasets, which are compared with state-of-the-art methods S-NeRF (~\cite{xie2022s}) and Zip-NeRF (~\cite{barron2023zip}). As shown in Fig.~\ref{fig: waymo}-~\ref{fig: nuscenesparanoma}, it is evident that both S-NeRF and Zip-NeRF exhibit color inconsistencies on the sides of the images, where they overlap with other cameras. In contrast, we have addressed this issue through layer-based color correction. Furthermore, we achieve improved rendering quality, such as clear contours, patterns, and text, and more accurate geometry by incorporating virtual warping and spatiotemporally constrained pose refinement. 

\end{document}